%% file: arxiv.tex
\documentclass{article}

\usepackage[utf8]{inputenc} %
\usepackage[T1]{fontenc}    %
\usepackage{microtype}
\usepackage{graphicx}

\usepackage{amsmath}
\usepackage{amssymb}
\usepackage{mathtools}
\usepackage{amsthm}

\usepackage{booktabs} %
\usepackage[aboveskip=2pt]{subcaption} %
\usepackage{wrapfig}

\usepackage[colorlinks=true,linkcolor=blue,allcolors=blue,urlcolor=blue,citecolor=blue]{hyperref}

\usepackage[accepted]{icml2023_dp4ml}

\usepackage[textsize=tiny]{todonotes}

\usepackage{tikz,pgfplots}
\usetikzlibrary{shapes,arrows,positioning,calc}
\usetikzlibrary{decorations.pathmorphing}
\usepackage[outline]{contour}
\pgfkeys{/pgf/number format/.cd,1000 sep={}}

\renewcommand{\paragraph}[1]{{\bf #1}~~}

\usepackage{tabularx}
\usepackage{array,multirow}
\usepackage{colortbl}
\newcommand{\PreserveBackslash}[1]{\let\temp=\\#1\let\\=\temp}
\newcolumntype{C}[1]{>{\PreserveBackslash\centering}p{#1}}

\usepackage{xspace}
\newcommand{\eg}{\textit{e.g.\@}\xspace}

\newcommand{\our}{\textsc{sfr}\xspace}

\usepackage{tikz}
\usepackage{pgfplots}
\usetikzlibrary{patterns}
\usetikzlibrary{decorations,backgrounds,arrows.meta,calc}
\usetikzlibrary{shapes,arrows,positioning}

\pgfplotsset{every axis/.append style={
		legend style={inner xsep=1pt, inner ysep=0.5pt, nodes={inner sep=1pt, text depth=0.1em},draw=none,fill=none}
}}

\usepackage{todonotes}
\usepackage{amsmath}
\usepackage{bm}
\usepackage{derivative}
\usepackage{wrapfig}

\usepackage{tikz,pgfplots}
\usepackage{subcaption}
\usetikzlibrary{}

\input{utils.tex}

\usepackage[capitalise,nameinlink]{cleveref}
\crefname{section}{Sec.}{Secs.}
\crefname{algorithm}{Alg.}{Algs.}
\crefname{appendix}{App.}{Apps.}
\crefname{definition}{Def.}{Defs.}
\newlength{\figurewidth}
\newlength{\figureheight}

\newcommand{\dataset}{\ensuremath{\mathcal{D}}}

\newcommand{\inputDomain}{\ensuremath{\mathbb{R}^{D}}}
\newcommand{\outputDomain}{\ensuremath{\mathbb{R}^{C}}}

\newcommand{\weights}{\ensuremath{\mathbf{w}}}

\usepackage{bm}
\newcommand{\mathbold}[1]{\bm{#1}}
\newcommand{\mbf}[1]{\mathbf{#1}}
\renewcommand{\mid}{\,|\,}

\newcommand{\MS}{\mbf{S}}

\newcommand{\MZ}{\mbf{Z}}

\newcommand{\MX}{\mbf{X}}

\newcommand{\MI}{\mbf{I}}

\newcommand{\T}{\top}
\newcommand{\vzeros}{\mbf{0}}

\newcommand{\valpha}[0]{\mathbold{\alpha}}

\newcommand{\vbeta}[0]{\mathbold{\beta}}
\newcommand{\MBeta}[0]{\mathbold{B}}

\newcommand{\diag}{\text{{diag}}}

\newcommand{\vm}{\mbf{m}}
\newcommand{\vz}{\mbf{z}}
\newcommand{\vf}{\mbf{f}}
\newcommand{\vu}{\mbf{u}}
\newcommand{\vx}{\mbf{x}}
\newcommand{\vy}{\mbf{y}}
\newcommand{\vw}{\mbf{w}}

\newcommand{\Jac}[2]{\mathcal{J}_{#1}(#2)}
\newcommand{\JacT}[2]{\mathcal{J}_{#1}^\top(#2)}

\newcommand{\MKzz}{\mbf{K}_{\mbf{z}\mbf{z}}}

\newcommand{\MKxx}{\mbf{K}_{\mbf{x}\mbf{x}}}

\newcommand{\vkzi}{\mbf{k}_{\mbf{z}i}}

\newcommand{\vkzs}{\mbf{k}_{\mbf{z}i}}
\newcommand{\vk}{\mbf{k}}

\definecolor{matplotlib-blue}{HTML}{1f77b4}

\newcommand{\myexpect}{\mathbb{E}}

\newcommand{\Norm}{\mathcal{N}}

\icmltitlerunning{Sparse Function-space Representation of Neural Networks}

\begin{document}

\twocolumn[
\icmltitle{Sparse Function-space Representation of Neural Networks}

\icmlsetsymbol{equal}{*}

\begin{icmlauthorlist}
\icmlauthor{Aidan Scannell}{equal,aalto,fcai}
\icmlauthor{Riccardo Mereu}{equal,aalto}
\icmlauthor{Paul Chang}{aalto}
\icmlauthor{Ella Tamir}{aalto}
\icmlauthor{Joni Pajarinen}{aalto}
\icmlauthor{Arno Solin}{aalto}
\end{icmlauthorlist}

\icmlaffiliation{aalto}{Aalto University, Espoo, Finland}
\icmlaffiliation{fcai}{Finnish Center for Artificial Intelligence}

\icmlcorrespondingauthor{Aidan Scannell}{aidan.scannell@aalto.fi}
\icmlkeywords{Machine Learning, ICML}

\vskip 0.3in
]

\printAffiliationsAndNotice{\icmlEqualContribution} %

\begin{abstract}
Deep neural networks (NNs) are known to lack uncertainty estimates and struggle to incorporate new data. We present a method that mitigates these issues by converting NNs from weight space to function space, via a dual parameterization. Importantly, the dual parameterization enables us to formulate a sparse representation that captures information from the entire data set. This offers a compact and principled way of capturing uncertainty and enables us to incorporate new data without retraining whilst maintaining predictive performance. We provide proof-of-concept demonstrations with the proposed approach for quantifying uncertainty in supervised learning on UCI benchmark tasks.
\end{abstract}

\section{Introduction}
Deep learning \citep{goodfellow2016deep} has become the cornerstone of contemporary artificial intelligence, proving remarkably effective in tackling supervised and unsupervised learning tasks in the {\em large data}, {\em offline}, and {\em gradient-based training} regime. Despite its success, gradient-based learning techniques exhibit limitations. Firstly, how can we efficiently quantify uncertainty without resorting to expensive and hard-to-interpret sampling in the model's weight space? Secondly, how to update the weights of an already trained model with new batches of data without compromising the performance on past data?
These questions become central when used for sequential learning,
such as continual learning \citep[CL,][]{parisi2019continual, de2021continual}, Bayesian optimization \citep[BO,][]{garnett_bayesoptbook_2022} and reinforcement learning  \citep[RL,][]{sutton2018reinforcement}.

\begin{figure}[t!]
  \centering\Large
  \pgfplotsset{axis on top,scale only axis,width=\figurewidth,height=\figureheight, ylabel near ticks,ylabel style={yshift=-2pt},y tick label style={rotate=90},legend style={nodes={scale=1., transform shape}},tick label style={font=\Large,scale=1}}
  \pgfplotsset{xlabel={Input, $x$},axis line style={rounded corners=2pt}}
  \setlength{\figurewidth}{.38\textwidth}
  \setlength{\figureheight}{\figurewidth}
  \def\inducing{\Huge Sparse inducing points}
  \begin{subfigure}[c]{.52\columnwidth}
    \raggedleft
    \pgfplotsset{ylabel={Output, $y$}}
    \input{./fig/regression-nn.tex}%
  \end{subfigure}
  \hfill
  \begin{subfigure}[c]{.02\columnwidth}
    \centering
    \tikz[overlay,remember picture]\node(p0){};
  \end{subfigure}
  \hfill
  \begin{subfigure}[c]{.4\columnwidth}
    \raggedleft
    \pgfplotsset{yticklabels={,,},ytick={\empty}}
    \input{./fig/regression-nn2svgp.tex}%
  \end{subfigure}
  \vspace*{-0.8em}
  \caption{\textbf{Regression example on an MLP with two hidden layers.} Left:~Predictions from the trained neural network. Right:~Our approach (\our) equips trained NNs with uncertainty estimates.} %
  \label{fig:teaser}
  \begin{tikzpicture}[remember picture,overlay]
    \tikzstyle{myarrow} = [draw=black!80, single arrow, minimum height=14mm, minimum width=2pt, single arrow head extend=4pt, fill=black!80, anchor=center, rotate=0, inner sep=5pt, rounded corners=1pt]
    \node[myarrow] (p0-arr) at ($(p0) + (0.2em,1.2em)$) {};
    \node[font=\scriptsize\sc,color=white] at (p0-arr) {\our};
  \end{tikzpicture}
  \vspace*{-1em}
\end{figure}
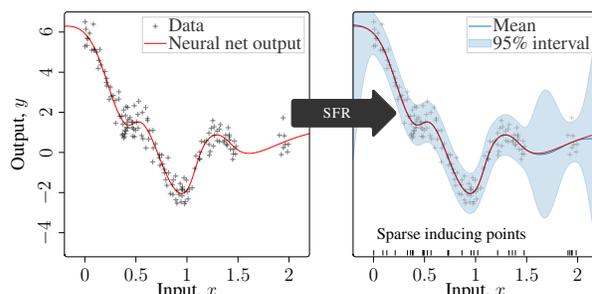

Recent techniques  \citep[\eg,][]{ritter2018kfac,khan2019approximate,daxberger2021laplace,fortuin2021bayesian,immer2021scalable} apply a Laplace
generalized Gauss-Newton (GGN) approximation to convert trained NNs into Bayesian neural networks (BNNs), that can provide uncertainty without sacrificing additional resources to training \citep{foong2019between}. Furthermore, the resultant weight-space posterior can be converted to function space as shown in \citet{khan2019approximate, immer2021improving}. The function-space representation allows for a principled mathematical approach for analyzing its behaviour \citep{cho2009kernel,meronen2020stationary}, performing probabilistic inference \citep{khan2019approximate}, and quantifying uncertainty \citep{foong2019between}. These methods rely on linearizing the NN and the resultant neural tangent kernel \citep[NTK,][]{jacot2018neural}.
The NN is characterized in function space by its first two moment functions, a mean function and covariance function (or kernel)---defining a Gaussian process \citep[GP,][]{rasmussen2006gaussian}. GPs provide a widely used probabilistic toolkit with principled uncertainty estimates.
Given an approximate inference technique, we demonstrate that NNs emit `dual' parameters which are the building blocks of a GP~\citep{csato2002sparse, adam2021dual, chang2023memory}.
In contrast to previous work that utilizes subsets of training data \citep{immer2021scalable}, our parameterization captures
information from {\em all} data points in a sparse representation, essential for scaling to deep learning data sets.
Importantly, the dual parameterization can be used to {\em (i)}~sparsify the GP without requiring further optimization
(\eg, variational inference) whilst capturing information from all data points, and {\em (ii)}~incorporate new data without retraining by conditioning on new data, in a computationally efficient manner.
The dual parameterization establishes a connection between NNs, GPs, and sparse approximations similar to sparse variational GPs~\citep{titsias2009variational}. We refer to our method as Sparse Function-space Representation (\our)---a sparse GP derived from a trained NN; see \cref{fig:teaser} for an example. %
\looseness-2

\section{Background}
\label{sec:methods}

We consider supervised learning with inputs $\vx_i \in \inputDomain$ and outputs $\vy_i \in \outputDomain$ (\eg, regression) or $\vy_{i} \in \{0,1\}^{C}$ (\eg, classification),
giving a data set $\dataset = \{(\vx_{i} , \vy_{i})\}_{i=1}^{N}$.
We introduce a  NN $f_\mathbf{w}: \inputDomain \to \outputDomain$ with weights $\weights \in \R^{P}$ and use a likelihood function $p(\vy_{i} \mid f_\mathbf{w}(\vx_{i}))$
to link the function values to the output $\vy_{i}$ (\eg, categorical for classification).
For notational conciseness, we stick to scalar outputs $y_{i}$. We denote the sets of all inputs
as $\MX = \{\vx_{i}\}_{i=1}^{N}$ and the set of all outputs $\vy = \{y_{i}\}_{i=1}^{N}$.

\textbf{BNNs}
In Bayesian deep learning, we place a prior over the weights $p(\vw)$ and aim to calculate the posterior over the weights given the data $p(\vw \mid \mathcal{D})$.
Given the weight posterior $p(\vw \mid \mathcal{D})$, we can make probabilistic predictions with
\begin{align} \label{eq-predictions}
  p_{\text{\sc BNN}}(y_{i} \mid \vx_{i}, \mathcal{D}) = \E_{p(\vw \mid \mathcal{D})} \left[ p(y_{i} \mid f_{\vw}(\vx_{i})) \right].
\end{align}
The posterior ${p(\vw \mid \dataset) \propto p(\vy \mid f_{\weights}(\MX)) \, p(\weights)}$ is generally not available in closed form
so we resort to approximations.
\textbf{MAP}
It is common to train NN weights $\vw$ to minimize the (regularized) empirical risk,
\begin{align} \label{eq-empirical-risk}
  \weights^{*} &=
                 \arg \min_{\weights} \mathcal{L}(\dataset,\weights) \nonumber \\
               &=
  \arg \min_{\weights} \textstyle\sum_{i=1}^{N} \underbrace{\ell(f_\weights(\mathbf{x}_{i}), y_i)}_{-\log p(y_{i} \mid f_{\vw}(\vx_{i}))} + \underbrace{\delta \mathcal{R}(\weights)}_{-\log p(\vw)}.
\end{align}
This objective corresponds to the log-joint distribution $\mathcal{L}(\mathcal{D},\vw)=p(\mathcal{D}, \vw)$
as the loss can be interpreted as a negative log-likelihood $\ell(f_\weights(\vx_{i}), y_i) = -\log(p(y_{i} \mid f_\weights(\vx_{i}))$
and the regularizer corresponds to a log prior over the weights $\delta\mathcal{R}(\weights) = -\log p(\vw)$.
For example, a weight decay regularizer $\mathcal{R}(\vw) = \frac{1}{2}\|\weights\|^{2}_2$ corresponds to a Gaussian prior over the weights $p(\vw) = \Norm(\vzeros, \delta^{-1} \MI)$, with
prior precision $\delta$.
As such, we can view \cref{eq-empirical-risk} as the maximum {\it a~posteriori} (MAP) solution.

\textbf{Laplace approximation}
The Laplace approximation \citet{mackayBayesian1992,daxberger2021laplace} builds upon this and approximates the weight posterior $p(\vw \mid \mathcal{D}) \approx q(\vw)= \mathcal{N}(\vw^{*} , \bm\Sigma)$
around the MAP weights ($\vw^{*}$) by setting the covariance to the Hessian of the posterior,
\begin{align} \label{eq:laplace-cov}
 \bm\Sigma = - \left[\nabla^{2}_{\vw\vw} \log p(\vw \mid \mathcal{D})  |_{\vw=\vw^{*}} \right]^{-1}.
\end{align}
Computing \cref{eq:laplace-cov} requires calculating the Hessian of the log-likelihood $\nabla^{2}_{\vw\vw} \log p(\vy \mid f_{\vw}(\MX))$ from \cref{eq-empirical-risk}.
As highlighted in \citet{immer2021improving}, computing this Hessian is often infeasible and in practice it is common to adopt the
generalized Gauss-Newton (GGN) approximation,
\begin{align}
 \nabla^{2}_{\vw\vw} \log p(\vy \mid f_{\vw}(\MX)) \approx \Jac{\vw}{\MX}^{\top} \nabla_{\vf\vf}^{2}\log(\vy\mid\vf) \Jac{\vw}{\MX}.
  \nonumber
\end{align}
where the Jacobian is given by $\Jac{\weights}{\vx} \coloneqq \left[ \nabla_\weights f_\weights(\vx)\right]^\top \in \R^{C \times P}$.
\citet{immer2021improving} highlighted that the GGN approximation corresponds to a local linearization of the NN,
$f_{\weights^{*}}^{\text{lin}}(\vx) = f_{\weights}(\vx) + \Jac{\weights^{*}}{\vx}(\weights-\weights^{*})$.
This suggests that predictions should be made with a generalized linear model (GLM), given by
\begin{align} \label{eq-glm}
  p_{\text{\sc GLM}}(y_{i} \mid \vx_{i}, \mathcal{D}) &= \E_{q(\vw)} \left[ p(y_{i} \mid f_{\vw^{*}}^{\text{lin}}(\vx_{i}))\right]. %
\end{align}

\textbf{Gaussian processes}
As Gaussian distributions remain tractable under linear transformations, we can convert the linear model from
weight space to function space \citep[see Ch.~2.1 in ][]{rasmussen2006gaussian}.
As such (and shown in \citet{immer2021improving}), the Bayesian GLM in \cref{eq-glm} has an equivalent GP formulation,
\begin{align} \label{eq-gp-pred-immer}
  p_{\text{\sc GP}}(y_{i} \mid \vx_{i}, \mathcal{D}) &= \E_{q(f_{i})} \left[ p(y_{i} \mid f_{i} )\right],
\end{align}
where $q(f_{i}) = \mathcal{N} \left( f_{\vw^{*}}(\vx_{i}), \sigma^{2}_{i} \right)$ is the GP's predictive posterior with variance,
\begin{align} \label{eq-gp-pred-immer-gp}
  \sigma^{2}_{i} &= k_{ii} - \vk_{\vx i}^\top ( \MKxx + \bm\Lambda^{-1})^{-1} \vk_{\vx i},
\end{align}
where the kernel $\kappa(\vx,\vx')=\frac{1}{\delta} \Jac{\weights^*}{\vx} \, \JacT{\weights^*}{\vx'}$ is the Neural Tangent Kernel \citep[NTK,][]{jacot2018neural}
and $f_i = f_\vw(\vx_i)$ is the function output at $\vx_i$.
The $ij$\textsuperscript{th} entry of matrix $\MKxx \in \R^{N \times N}$ is $\kappa(\vx_i,\vx_j)$, $\vk_{\vx i}$ is a vector where
each $j$\textsuperscript{th} element is $\kappa(\vx_i, \vx_j)$, $k_{ii} = \kappa(\vx_i, \vx_i)$ and $\bm\Lambda = - \nabla^2_{\vf \vf}\log p(\vy \mid \vf)$ can be interpreted as per-input noise.

\textbf{Sparse GPs}
The GP formulation from \citet{immer2021improving} (\cref{eq-gp-pred-immer})
requires inverting an $N\times N$ matrix which has complexity $O(N^3)$.
This limits its applicability to large data sets, which are common in deep learning.
Sparse GPs reduce the computational complexity by representing the GP as a low-rank approximation at a set of inducing
inputs $\MZ = \{\vz_{j} \}_{j=1}^{M} \in \R^{M \times D}$ with corresponding inducing variables $\vu = f(\MZ)$ \citep[see][for an early overview]{quinonero2005unifying}.
The approach by \citet{titsias2009variational} (also used in the DTC approximation),
defines the marginal predictive distribution as $q_{\vu}(f_i)  = \int p(f_i  \mid \vu) \, q(\vu) \, \mathrm{d}\vu$ where
the variational distribution is parameterized as $q(\vu) = \mathcal{N}\left(\vu \mid \vm, \MS \right)$ (as in \citet{titsias2009variational,hensman2013gaussian}).
Assuming a zero mean function, the sparse GP predictive posterior is
\begin{subequations}  \label{eq-svgp-pred}
\begin{align}
  \myexpect_{q_{\vu}(f_i)}[f_i] &= \vk_{\vz i}^{\top}\MKzz^{-1}\vm, \\
  \mathrm{Var}_{q_{\vu}(f_i)}[f_i] &= k_{ii} - \vk_{\vz i}^\top ( \MKzz^{-1} -  \MKzz^{-1}\MS  \MKzz^{-1}) \vk_{\vz i},  \nonumber
\end{align}
\end{subequations}
where $\MKzz$ and $\vkzs$ are defined similarly to $\MKxx$ and $\vk_{\vx i}$ but over the inducing points $\MZ$.%
Note that the parameters of the variational posterior ($\vm$ and $\MS$) are usually obtained via variational
inference, which requires further optimization.%

Not only does the GP formulation from \citet{immer2021improving} (shown in \cref{eq-gp-pred-immer}) struggle to scale to large
data sets but it also cannot incorporate new data by conditioning on it.
This is because its posterior mean is the NN  $f_{\vw^*}(\vx)$.
Our method overcomes both of these limitations via a dual parameterization.
Importantly, our dual parameterization enables us to {\em (i)}~sparsify the GP without further optimization, and {\em (ii)}~incorporate new data without retraining.\looseness-1 %

\section{\our: Sparse function-space representation of NNs}\label{sec:sfr}
In this section, we present our method, named \our, which converts a trained NN into a GP.
\our is built upon a dual parameterization of the GP posterior.
Early work by \citet{csato2002sparse} showed that Expectation Propagation gives rise to a dual parameterization.
More recent work by \citet{adam2021dual,chang2023memory} showed a different dual parameterization arising from the evidence lower bound for sparse variational GPs \citep{hensman2013gaussian,titsias2009variational}.
The dual parameterization from \citet{adam2021dual,chang2023memory}, consists of parameters $\valpha$ and $\vbeta$, which gives rise to the predictive posterior,
\begin{subequations}  \label{eq:gp_pred}
\begin{align}
  \myexpect_{p(f_i \mid\vy)}[f_i] &= \vk_{\vx i}^\top \valpha, \\
  \mathrm{Var}_{p(f_i \mid \vy)}[f_i] &= k_{ii} - \vk_{\vx i}^\top ( \MKxx + \diag(\vbeta)^{-1})^{-1} \vk_{\vx i}.  \nonumber
\end{align}
\end{subequations}
\cref{eq:gp_pred} states that the first two moments of the resultant posterior process (which may not be a GP), can be parameterized via the dual
parameters $\valpha, \vbeta \in \R^{N}$,
defined as \looseness-1
\begin{subequations}
\label{eq:dual_param}
\begin{align}
  \alpha_i &\coloneqq \myexpect_{p(\vw \mid \vy)}[\nabla_{f}\log p(y_i \mid f) |_{f=f_i}], \\
  \beta_i &\coloneqq - \myexpect_{p(\vw \mid \vy)}[\nabla^2_{f f}\log p(y_i \mid f_i) |_{f=f_i}].
  \end{align}
\end{subequations}
\cref{eq:dual_param} holds for generic likelihoods and involves no approximations since the expectation is under the
exact posterior, given that the model can be expressed in a kernel formulation.
\cref{eq:gp_pred} and \cref{eq:dual_param} highlight that the approximate inference technique, usually viewed as a posterior approximation,
can alternatively be interpreted as an approximation of the expectation of loss (likelihood) gradients.

\paragraph{Dual parameters from NN}
Given that we use a Laplace approximation of the NN, we remove the expectation over the posterior \citep[see Ch.~3.4.1 in][for derivation]{rasmussen2006gaussian} and our dual parameters are given by
\begin{subequations}
\label{eq:dual_param_laplace}
\begin{align}
  \hat{\alpha}_i &\coloneqq \nabla_{f}\log p(y_i \mid f) |_{f=f_i} , \\
  \hat{\beta}_i &\coloneqq - \nabla^2_{ff}\log p(y_i \mid f) |_{f=f_i}.
  \end{align}
\end{subequations}
Substituting \cref{eq:dual_param_laplace} into \cref{eq:gp_pred}, we obtain our GP based on the trained NN.
Making predictions with \cref{eq:gp_pred} costs $O(N^3)$, which limits its applicability on large data sets.

\begin{table*}[t!]
  \centering\scriptsize
  \caption{Comparisons on UCI data with negative log predictive density (NLPD\textcolor{gray}{\footnotesize$\pm$std}, lower better). \our (with $M=20\% \text{ of } N$) is on par with the Laplace approximation (BNN/GLM) and outperforms the GP subset when the prior precision $(\delta)$ is tuned (right). Interestingly, when the prior precision ($\delta$) is not tuned (left), \our outperforms all other methods. The GP subset uses the same inducing points as \our.} %
	\label{tbl:uci}
	\setlength{\tabcolsep}{1.8pt}

	\newcommand{\val}[2]{%
		$#1$\textcolor{gray}{${\pm}#2$}
	}

	\input{tables/workshop_uci_table.tex}
\end{table*}

\begin{figure*}[t]
  \centering\small
  \setlength{\figurewidth}{.27\textwidth}
  \setlength{\figureheight}{\figurewidth}
  \pgfplotsset{axis on top,scale only axis,y tick label style={rotate=90}, x tick label style={font=\Large},y tick label style={font=\Large},title style={yshift=-4pt,font=\LARGE}, y label style={font=\Large},x label style={font=\LARGE},grid=major, width=\figurewidth, height=\figureheight}
  \pgfplotsset{grid style={line width=.1pt, draw=gray!10,dashed}}
  \pgfplotsset{xlabel={$M$ as \% of $N$},ylabel style={yshift=-5pt},xlabel style={yshift=-5pt}}
  \begin{minipage}[t]{.175\textwidth}
    \raggedleft
    \pgfplotsset{ylabel=NLPD}
    \input{./fig/australian.tex}
  \end{minipage}
  \hfill
  \begin{minipage}[t]{.155\textwidth}
    \raggedleft
    \input{./fig/glass.tex}
  \end{minipage}
  \hfill
  \begin{minipage}[t]{.155\textwidth}
    \raggedleft
    \input{./fig/vehicle.tex}
  \end{minipage}
  \hfill
  \begin{minipage}[t]{.155\textwidth}
    \raggedleft
    \input{./fig/waveform.tex}
  \end{minipage}
  \hfill
  \begin{minipage}[t]{.155\textwidth}
    \raggedleft
    \input{./fig/digits.tex}
  \end{minipage}
  \hfill
  \begin{minipage}[t]{.155\textwidth}
    \raggedleft
    \input{./fig/satellite.tex}
  \end{minipage}\\[-1em]
  \definecolor{steelblue31119180}{RGB}{31,119,180}
  \definecolor{darkorange25512714}{RGB}{255,127,14}
  \newcommand{\myline}[1]{\protect\tikz[baseline=-.5ex,line width=1.6pt]\protect\draw[draw=#1](0,0)--(1.2em,0);}
  \caption{Comparison of convergence in terms of number of inducing points $M$ in NLPD (mean\textcolor{gray}{\footnotesize$\pm$std} over 5 seeds) on UCI classification tasks: \our (\myline{steelblue31119180}) vs. GP subset (\myline{darkorange25512714}). Our \our converges fast for all cases showing clear benefits of its ability to summarize all the data onto a sparse set of inducing points. The number of inducing points $M$ are specified as a percentage of the number of training data $N$.}
  \label{fig:uci}
\end{figure*}
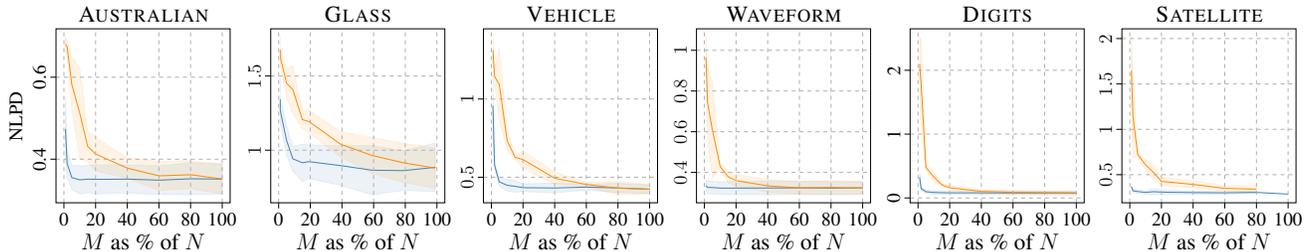

\paragraph{Sparsification via dual parameters}
\label{sec:sparse-dual-gp}
Given that we have computed the dual parameters derived from our NN predictions and a kernel function, we could essentially employ any sparsification method \citep{quinonero2005unifying}.
In this work, we opt for the approach suggested by \citet{titsias2009variational,hensman2013gaussian},
which is shown in \cref{eq-svgp-pred}.
However, instead of parameterizing the variational distribution as $q(\vu) = \mathcal{N}\left(\vu \mid \vm, \MS \right)$,
we follow insights from \citet{adam2021dual} that the posterior under this model bears a structure akin to \cref{eq:gp_pred}.
As such, we project the dual parameters onto the inducing points giving sparse dual parameters.
Using this sparse definition of the dual parameters, our sparse GP posterior is given by
\begin{subequations} \label{eq:dual_sparse_post}
\begin{align}
   \myexpect_{q_{\vu}(\vf)}[f_i] &= \vkzs^{\T} \MKzz^{-1} \valpha_{\vu}, \\
   \textrm{Var}_{q_{\vu}(\vf)}[f_i]  &= k_{ii} - \vkzs^\top [\MKzz^{-1} - (\MKzz + \MBeta_{\vu})^{-1} ]\vkzs, \nonumber
\end{align}
\end{subequations}
with sparse dual parameters,
\begin{equation} \textstyle
  \valpha_{\vu}  =  \sum_{i=1}^N  \vkzi \, \hat{\alpha}_{i}
  \quad \text{and} \quad
  \MBeta_{\vu} =  \sum_{i=1}^N \vkzi \,\hat{\beta}_{i} \, \vkzi^{\T} .
\label{eq:dual_sparse}
\end{equation}
Note that the sparse dual parameters are now a sum over \emph{all data points}, with $\valpha_{\vu} \in \R^{M}$ and $\MBeta_{\vu} \in \R^{M  \times M}$.
Contrasting \cref{eq:dual_sparse_post} and \cref{eq:gp_pred}, we can see that the computational complexity went from $\mathcal{O}(N^3)$ to $\mathcal{O}(M^3)$, with $M \ll N$.
Crucially, given the structure of our probabilistic model, our sparse dual parameters \cref{eq:dual_sparse} are a compact representation of the full model projected using the kernel.

We highlight that \our differs from the GP subset from \citet{immer2021improving}, which results in several advantages.
First, \our leverages a dual parameterization to construct a sparse GP, which captures information from the entire data set.
In contrast, \citet{immer2021improving} utilize a data subset which ignores information from the rest of the data set.
Second, \our makes predictions with the GP mean, whereas \citet{immer2021improving} center predictions around NN predictions $f_{\vw^*}(\cdot)$.
As such, \our can incorporate new data without retraining whilst the GP subset cannot.

To the best of our knowledge, we are the first to formulate a dual GP from a NN.
Our dual parameterization has two main benefits: {\em (i)}~it enables us to construct a sparse GP without requiring us to perform further optimization, and {\em (ii)}~it enables us to incorporate new data without retraining by conditioning on new data (using \cref{eq:dual_sparse}).

\section{Experiments}
\label{sec:experiments}
We evaluate \our on eight UCI~\citep{UCI} binary and multi-class classification tasks.
Following \citet{immer2021improving}, we used a two-layer MLP with width 50, tanh activation functions and a
$70\%\ (\text{train}):15\%\ (\text{validation}) :15\%\ (\text{test})$ data split.
We trained the NN using Adam \cite{adam} with a learning rate of $10^{-4}$ and a batch size of $128$.
Training was stopped when the validation negative log predictive density (NLPD) stopped decreasing after 1000 steps.
The checkpoint with the lowest NLPD was the NN MAP.
Each experiment was run for $5$ seeds.\looseness-1

We first compare \our to the Laplace approximation when making predictions with {\em (i)}~the nonlinear NN (BNN) and {\em (ii)}~the generalised linear model (GLM) in \cref{eq-glm}.
We use the Laplace PyTorch library \citep{daxberger2021laplace} with the full Hessian. %
It is common practice to tune the prior precision $\delta$ after training the NN.
That is, find a prior precision $\delta_{\text{tune}}$ which has a better NLPD on the validation set than the $\delta$ used to train the NN.
Note that this leads to a different prior precision being used for training and inference.
We highlighted that technically this invalidates the Laplace/SFR methods as the NN weights are no longer the MAP weights.
Nevertheless, we report results when the prior precision $\delta$ is tuned (\cref{tbl:uci} right) and is not tuned (\cref{tbl:uci} left).
When tuning the prior precision $\delta$, \our matches the performance of the Laplace methods (BNN/GLM).
Interestingly, \our also  performs well without tuning the prior precision $\delta$, unlike the Laplace methods.

As \our captures information from all of the training data in the inducing points, we compare \our to making GP predictions with a subset of the training data (GP subset).
To make the comparison fair, we use \our's inducing inputs $\vz$ as the subset.
\cref{tbl:uci} shows that \our is able to summarize the full data set more effectively than the GP subset method as it maintains predictive performance
whilst using fewer inducing points.
This is further illustrated in \cref{fig:uci}, which shows that as the number of inducing points is lowered from $M=100\%$ of $N$ to $M=1\%$ of $N$,
\our is able to maintain a much better NLPD.
This demonstrates that \our  captures information from the entire data set and as a result can use fewer inducing points than the GP subset.

\section{Discussion and conclusion}
\label{sec:conclusion}
We introduced \our, a novel approach for representing NNs in function space.
It leverages a dual parameterization to obtain a low-rank (sparse) approximation that captures information from the entire data set.
We showcased \our's uncertainty quantification in UCI benchmark classification tasks (\cref{tbl:uci}).
Unlike the Laplace approximation, \our achieves good predictive performance without tuning the prior precision $\delta$.
This is interesting and we believe it stays more true to Bayesian inference.%
In practical terms, \our serves a role similar to a sparse GP. However, unlike a conventional GP, it does not provide a straightforward method for specifying the prior.%
This limitation can be addressed indirectly: the architecture of the NN and the choice of activation functions can be used to implicitly specify the prior assumptions.
It is important to note that \our linearizes the NN around the MAP weights $\weights^{*}$, resulting in the function-space prior
(and posterior) being a locally linear approximation of the NN.\looseness-1

\clearpage

\bibliographystyle{icml2023}

\newpage
\appendix
\onecolumn

\end{document}

%% file: utils.tex
\usepackage{mathtools}

\DeclareMathOperator{\R}{\mathbb{R}}
\DeclareMathOperator{\E}{\mathbb{E}}

%% file: fig/regression-nn.tex
\begin{tikzpicture}[scale=0.5]

\definecolor{darkgray176}{RGB}{176,176,176}
\definecolor{lightgray204}{RGB}{204,204,204}

\begin{axis}[
height=\figureheight,
legend cell align={left},
legend style={fill opacity=0.8, draw opacity=1, text opacity=1, draw=lightgray204},
tick align=outside,
tick pos=left,
width=\figurewidth,
x grid style={darkgray176},
xmin=-0.2, xmax=2.2,
xtick style={color=black},
y grid style={darkgray176},
ymin=-5.2, ymax=7,
ytick style={color=black}
]
\addplot [draw=black, fill=black, mark=+, only marks, opacity=0.5]
table{%
x  y
0.59907633700711 0.223154562086841
0.34361536741137 1.80796222965741
1.21549153760962 0.967955764303031
1.39127490332365 0.346027494892729
0.527962441193321 2.55278258834587
0.0333065151243783 5.67346791281363
1.41384715629159 0.881468575721158
0.663171303916053 -0.472049787519171
1.00205910730797 -1.96751300267146
0.420018112144075 1.21728173287471
0.631458340576862 1.08436032794776
0.729874319093148 -0.101521243817772
0.147377188445248 5.13582753705874
0.403831786198692 2.27647279843891
0.478013434900342 1.17179814470945
0.595830829634752 2.23305061724589
0.955601779278568 -1.72905380014215
0.0839045702835102 6.3929614602345
1.45472441165539 -0.0760227450398461
0.278479698750005 2.84103829741965
0.380362104273414 2.11325866978181
0.464499503069764 1.2399605663803
0.793107714094262 -1.35173014471992
0.977768289229485 -2.55927294171402
0.569210172057486 0.68494103913613
0.36980976303721 0.921314385700647
0.0214569812169767 5.68468381545469
0.424236769219515 1.63566013099165
1.37938987893611 0.868125317911011
0.137047115400012 4.58721934956631
1.08761751839489 -0.172567506567352
0.419817741770634 2.24046262513834
1.4262841184116 0.159812522509929
1.41564972631546 0.78558633647277
0.495989002819549 1.10158598995442
0.995026761218884 -1.37546551579343
1.12639463876016 -0.334794924441981
1.07766860579985 -0.792376161207397
0.413701619771477 1.4913623618045
0.942986552597874 -2.4722425346432
0.763519126454151 -0.583688218859911
0.130904366804652 4.18156576167409
0.0454384945865041 5.94112904944771
0.522204951265 1.21564815606939
0.332942022904396 2.65994126760993
1.40363369449562 1.23353940603206
0.779310595942776 -1.02035508529735
1.29407529214362 1.06264876539534
0.556033388135621 0.837778673324579
0.484633763104316 2.32967153874662
0.0920321107558388 5.09987558004654
0.56325131198675 1.98910435881925
1.45678884132294 0.154899001388011
0.408702868409535 1.25289558714472
0.458097584655798 1.52690656054213
1.46582759713057 0.56502069721856
0.507184083096661 1.52564677260393
1.36046218960706 0.507926246238466
0.921764573531882 -2.09142805291535
0.238639274847773 2.6411155956647
0.362559658595395 1.79727559711587
1.21589080021465 0.400000674361935
0.933585205353682 -1.55461336837348
0.982715141501147 -2.45533624593891
0.401490860442604 1.05987192830147
0.156059199998835 4.36370190980214
0.891378994342085 -2.36391299232751
0.388158878064094 0.788154469107815
0.667188047358641 1.10183787343592
0.625170599258625 0.6586301610886
0.957021857615636 -1.85051954010725
1.02156844318979 -1.83957136491289
1.25724919181647 0.322518415235646
0.00266557165722991 6.50557346836786
1.41816728826336 0.629351904749414
1.31945988198246 0.544524900201801
1.25294153993982 1.58846644417793
1.44809592439576 -0.343645665284646
1.12607739077688 0.0198156968068374
0.251530997239499 3.26092402478648
0.729180476405212 -0.261848785696243
0.902909475880636 -1.88887529622365
0.815297425524211 -0.263213218540796
0.399596820896549 1.8350776647408
1.19233932331138 0.594739208851601
0.212701803984358 3.69662943902046
0.48579647389144 0.781131500953395
0.754043399884807 -0.120330706281183
0.299033951960917 2.59001615301901
0.471011920030608 0.803031316713116
1.23848755238611 1.20117380925507
0.0927035953462867 5.14006372858808
0.803635270801682 -1.64573382353524
0.971532479828419 -1.26141546874724
0.866704965181163 -1.37016791118963
1.36729265920189 0.597849962429656
0.519752323523261 1.89624294678323
0.0191107582149252 5.33379579237503
0.196986770599228 4.03647401377799
0.952624784132827 -2.16148044060619
0.0015044650302265 5.2886136068202
0.902037550168227 -2.52858794998143
1.22749030238651 0.0464677514177437
1.21792105205884 -0.141830633035098
0.0102856385216306 6.14133785166822
0.863182937449097 -2.1011551265248
0.818564275101864 -1.01534185724427
0.302635244152368 1.46210778674698
0.372879666668875 1.40476340640872
0.889792662887503 -1.17727077360789
1.29641079289544 0.586535140325398
1.3272574482414 1.3843767453822
0.214325965580719 3.31884710039265
0.614137818457737 1.59184679724417
0.368321423663743 0.845629772651128
0.380933232697981 0.723731284253583
0.976159176116351 -1.93326574524374
0.857441599514585 -1.60433853663788
0.83708880960736 -2.04818343189899
1.47579759938635 -0.0805765439070799
0.442438891644387 1.73835028751105
1.38953493744193 0.417026784527823
1.14080627628873 0.550112184891199
1.48114359376883 0.298669829133318
0.736597976181201 -0.643077812882977
0.394694513831174 0.936428225891781
1.26908189826216 0.858430359609534
0.0672098890730217 5.09900003215748
0.333851467177716 3.01981294882107
0.945969964783799 -1.83629872793283
0.692843988254517 0.433249498499454
1.26598975868233 1.28216661393988
0.391972523919673 1.09693715415414
0.46917601169147 1.50263243659036
1.01098339289849 -2.31817336136178
0.777738706583271 -0.838717602177543
1.12209237980542 -0.319615415703505
1.17055379105428 0.512465979308082
0.258836084652896 2.78755665557385
0.61396534548555 1.69052927515609
1.31770140735241 1.16502121252192
0.116000397751632 4.99444756152836
0.210700698077648 3.49050340807256
1.05114996874651 -1.76278149811195
0.670454848751649 0.224613670765436
1.92406486757113 -0.108425407068489
1.94209538616793 1.7264746642509
1.99393444419407 0.403999409461401
1.95439877208446 0.827632344838344
1.98473609767015 0.0244378792655042
1.9054919129869 0.855315202370139
1.94542261961925 0.240943330223005
1.96246848086935 0.398027059259599
};
\addlegendentry{Data}
\addplot [semithick, red]
table {%
-0.2 6.2857973514481
-0.187939698492462 6.28987893254762
-0.175879396984925 6.29181588372522
-0.163819095477387 6.29143317385438
-0.151758793969849 6.28854013004549
-0.139698492462312 6.28292926239899
-0.127638190954774 6.27437510005683
-0.115577889447236 6.26263307579898
-0.103517587939698 6.24743850896844
-0.0914572864321608 6.22850575195502
-0.0793969849246231 6.20552758417457
-0.0673366834170854 6.17817495967054
-0.0552763819095477 6.14609724012402
-0.0432160804020101 6.10892307376503
-0.0311557788944724 6.06626211137595
-0.0190954773869347 6.01770778129061
-0.00703517587939698 5.96284137276835
0.00502512562814073 5.90123769647965
0.0170854271356784 5.83247259523151
0.0291457286432161 5.75613255854249
0.0412060301507538 5.67182664040638
0.0532663316582915 5.5792007786667
0.0653266331658292 5.47795445565633
0.0773869346733668 5.36785941553933
0.0894472361809046 5.24877986427611
0.101507537688442 5.12069323625788
0.11356783919598 4.9837102481237
0.125628140703518 4.83809262672679
0.137688442211055 4.68426666625203
0.149748743718593 4.52283072306446
0.161809045226131 4.35455497673142
0.173869346733668 4.18037232676687
0.185929648241206 4.00136016107162
0.197989949748744 3.81871385699869
0.210050251256281 3.63371411567772
0.222110552763819 3.44769137777358
0.234170854271357 3.26199138801039
0.246231155778894 3.0779462474231
0.258291457286432 2.89685486141262
0.27035175879397 2.71997550441124
0.282412060301508 2.54853134759842
0.294472361809045 2.38372743454599
0.306532663316583 2.2267750630501
0.318592964824121 2.07891724952624
0.330653266331658 1.94144733602477
0.342713567839196 1.81571216298775
0.354773869346734 1.70309166873519
0.366834170854271 1.60494816112118
0.378894472361809 1.52254067772778
0.390954773869347 1.45690296965451
0.403015075376884 1.40868845690316
0.415075376884422 1.37799311124833
0.42713567839196 1.36417806680222
0.439195979899497 1.36572605810096
0.451256281407035 1.38017446012216
0.463316582914573 1.40416536568561
0.475376884422111 1.43363406884371
0.487437185929648 1.46412297935564
0.499497487437186 1.49116970386928
0.511557788944724 1.51069284803595
0.523618090452261 1.51929979398149
0.535678391959799 1.51446714403481
0.547738693467337 1.49458367539806
0.559798994974875 1.45888030004499
0.571859296482412 1.40729020308166
0.58391959798995 1.3402834770427
0.595979899497487 1.2587099228498
0.608040201005025 1.16366898121746
0.620100502512563 1.05641276278727
0.632160804020101 0.938279469283438
0.644221105527638 0.81065027839526
0.656281407035176 0.674921907157222
0.668341708542714 0.53248820908157
0.680402010050251 0.384726155567839
0.692462311557789 0.232983617124391
0.704522613065327 0.0785680401076337
0.716582914572864 -0.0772637885233033
0.728643216080402 -0.233314196261137
0.74070351758794 -0.388450594830113
0.752763819095478 -0.541608554626875
0.764824120603015 -0.691791981866757
0.776884422110553 -0.838070751640898
0.788944723618091 -0.979576344982615
0.801005025125628 -1.11549603458366
0.813065326633166 -1.24506598531361
0.825125628140704 -1.36756334671807
0.837185929648241 -1.48229709669328
0.849246231155779 -1.58859712629165
0.861306532663317 -1.68580089703069
0.873366834170854 -1.773236998834
0.885427135678392 -1.85020512459456
0.89748743718593 -1.915952398347
0.909547738693467 -1.96964671375817
0.921608040201005 -2.01034885974521
0.933668341708543 -2.03698686496384
0.945728643216081 -2.0483383129386
0.957788944723618 -2.04302939259167
0.969849246231156 -2.01956288769164
0.981909547738694 -1.97639030730567
0.993969849246231 -1.91204409703886
1.00603015075377 -1.82534140335192
1.01809045226131 -1.7156576560425
1.03015075376884 -1.5832440846665
1.04221105527638 -1.42953085004441
1.05427135678392 -1.25732787946063
1.06633165829146 -1.07082794450782
1.078391959799 -0.875349345818019
1.09045226130653 -0.676830101549934
1.10251256281407 -0.481174989533318
1.11457286432161 -0.293615776282351
1.12663316582915 -0.118239365629348
1.13869346733668 0.0422281830821599
1.15075376884422 0.186383598074527
1.16281407035176 0.313924893349873
1.1748743718593 0.425342584611239
1.18693467336683 0.521607339752872
1.19899497487437 0.603906758694623
1.21105527638191 0.673452590391777
1.22311557788945 0.731357540851534
1.23517587939698 0.778569491602677
1.24723618090452 0.815847807322585
1.25929648241206 0.843767921655468
1.2713567839196 0.862743743355362
1.28341708542714 0.873060917720606
1.29547738693467 0.874916788169473
1.30753768844221 0.868464737975285
1.31959798994975 0.853861424167359
1.33165829145729 0.831315338883324
1.34371859296482 0.801134330340053
1.35577889447236 0.763768482133921
1.3678391959799 0.719843528247159
1.37989949748744 0.670179340507215
1.39195979899498 0.615788538744166
1.40402010050251 0.557852295893885
1.41608040201005 0.497673827854153
1.42814070351759 0.436614167370649
1.44020100502513 0.376018470128547
1.45226130653266 0.317143106662216
1.4643216080402 0.261093458268937
1.47638190954774 0.208779774388998
1.48844221105528 0.160894532365159
1.50050251256281 0.11791068459771
1.51256281407035 0.0800970364996313
1.52462311557789 0.0475453189557888
1.53668341708543 0.0202032889778795
1.54874371859296 -0.00209098424899307
1.5608040201005 -0.0195771400316618
1.57286432160804 -0.032544923219021
1.58492462311558 -0.0413118230085809
1.59698492462312 -0.0462059543733475
1.60904522613065 -0.0475536584522557
1.62110552763819 -0.045671049320416
1.63316582914573 -0.0408586771824737
1.64522613065327 -0.0333985352335383
1.6572864321608 -0.0235527505120653
1.66934673366834 -0.0115634290159743
1.68140703517588 0.00234675048540267
1.69346733668342 0.0179734921058688
1.70552763819095 0.0351295990490659
1.71758793969849 0.0536435941307043
1.72964824120603 0.073358338275725
1.74170854271357 0.0941297045827736
1.75376884422111 0.115825340404262
1.76582914572864 0.138323532947578
1.77788944723618 0.161512182936473
1.78994974874372 0.185287884140754
1.80201005025126 0.209555102768635
1.81407035175879 0.234225448846351
1.82613065326633 0.259217031081335
1.83819095477387 0.284453886822803
1.85025125628141 0.309865479256349
1.86231155778894 0.335386254670267
1.87437185929648 0.360955253367883
1.88643216080402 0.38651576848721
1.89849246231156 0.41201504758538
1.9105527638191 0.437404032335556
1.92261306532663 0.462637132073363
1.93467336683417 0.487672027233049
1.94673366834171 0.512469498951547
1.95879396984925 0.536993281313269
1.97085427135678 0.561209932880848
1.98291457286432 0.585088724324853
1.99497487437186 0.608601539142692
2.0070351758794 0.631722784652697
2.01909547738693 0.654429310669023
2.03115577889447 0.676700333507167
2.04321608040201 0.698517363236846
2.05527638190955 0.719864132383363
2.06733668341709 0.740726524574086
2.07939698492462 0.761092501925605
2.09145728643216 0.780952030261601
2.1035175879397 0.800297001533838
2.11557788944724 0.819121153082388
2.12763819095477 0.837419983610359
2.13969849246231 0.855190665958855
2.15175879396985 0.872431956946856
2.16381909547739 0.889144104686578
2.17587939698492 0.905328753897681
2.18793969849246 0.920988849824405
2.2 0.936128541410374
};
\addlegendentry{Neural net output}
\end{axis}

\end{tikzpicture}

%% file: fig/regression-nn2svgp.tex
\begin{tikzpicture}[scale=0.5]

\definecolor{darkgray176}{RGB}{176,176,176}
\definecolor{lightgray204}{RGB}{204,204,204}
\definecolor{steelblue31119180}{RGB}{31,119,180}

\begin{axis}[
height=\figureheight,
legend cell align={left},
legend style={fill opacity=0.8, draw opacity=1, text opacity=1, draw=lightgray204},
tick align=outside,
tick pos=left,
width=\figurewidth,
x grid style={darkgray176},
xmin=-0.2, xmax=2.2,
xtick style={color=black},
y grid style={darkgray176},
ymin=-5.2, ymax=7,
ytick style={color=black}
]
\addplot [draw=black, fill=black, forget plot, mark=+, only marks, opacity=0.2]
table{%
x  y
0.59907633700711 0.223154562086841
0.34361536741137 1.80796222965741
1.21549153760962 0.967955764303031
1.39127490332365 0.346027494892729
0.527962441193321 2.55278258834587
0.0333065151243783 5.67346791281363
1.41384715629159 0.881468575721158
0.663171303916053 -0.472049787519171
1.00205910730797 -1.96751300267146
0.420018112144075 1.21728173287471
0.631458340576862 1.08436032794776
0.729874319093148 -0.101521243817772
0.147377188445248 5.13582753705874
0.403831786198692 2.27647279843891
0.478013434900342 1.17179814470945
0.595830829634752 2.23305061724589
0.955601779278568 -1.72905380014215
0.0839045702835102 6.3929614602345
1.45472441165539 -0.0760227450398461
0.278479698750005 2.84103829741965
0.380362104273414 2.11325866978181
0.464499503069764 1.2399605663803
0.793107714094262 -1.35173014471992
0.977768289229485 -2.55927294171402
0.569210172057486 0.68494103913613
0.36980976303721 0.921314385700647
0.0214569812169767 5.68468381545469
0.424236769219515 1.63566013099165
1.37938987893611 0.868125317911011
0.137047115400012 4.58721934956631
1.08761751839489 -0.172567506567352
0.419817741770634 2.24046262513834
1.4262841184116 0.159812522509929
1.41564972631546 0.78558633647277
0.495989002819549 1.10158598995442
0.995026761218884 -1.37546551579343
1.12639463876016 -0.334794924441981
1.07766860579985 -0.792376161207397
0.413701619771477 1.4913623618045
0.942986552597874 -2.4722425346432
0.763519126454151 -0.583688218859911
0.130904366804652 4.18156576167409
0.0454384945865041 5.94112904944771
0.522204951265 1.21564815606939
0.332942022904396 2.65994126760993
1.40363369449562 1.23353940603206
0.779310595942776 -1.02035508529735
1.29407529214362 1.06264876539534
0.556033388135621 0.837778673324579
0.484633763104316 2.32967153874662
0.0920321107558388 5.09987558004654
0.56325131198675 1.98910435881925
1.45678884132294 0.154899001388011
0.408702868409535 1.25289558714472
0.458097584655798 1.52690656054213
1.46582759713057 0.56502069721856
0.507184083096661 1.52564677260393
1.36046218960706 0.507926246238466
0.921764573531882 -2.09142805291535
0.238639274847773 2.6411155956647
0.362559658595395 1.79727559711587
1.21589080021465 0.400000674361935
0.933585205353682 -1.55461336837348
0.982715141501147 -2.45533624593891
0.401490860442604 1.05987192830147
0.156059199998835 4.36370190980214
0.891378994342085 -2.36391299232751
0.388158878064094 0.788154469107815
0.667188047358641 1.10183787343592
0.625170599258625 0.6586301610886
0.957021857615636 -1.85051954010725
1.02156844318979 -1.83957136491289
1.25724919181647 0.322518415235646
0.00266557165722991 6.50557346836786
1.41816728826336 0.629351904749414
1.31945988198246 0.544524900201801
1.25294153993982 1.58846644417793
1.44809592439576 -0.343645665284646
1.12607739077688 0.0198156968068374
0.251530997239499 3.26092402478648
0.729180476405212 -0.261848785696243
0.902909475880636 -1.88887529622365
0.815297425524211 -0.263213218540796
0.399596820896549 1.8350776647408
1.19233932331138 0.594739208851601
0.212701803984358 3.69662943902046
0.48579647389144 0.781131500953395
0.754043399884807 -0.120330706281183
0.299033951960917 2.59001615301901
0.471011920030608 0.803031316713116
1.23848755238611 1.20117380925507
0.0927035953462867 5.14006372858808
0.803635270801682 -1.64573382353524
0.971532479828419 -1.26141546874724
0.866704965181163 -1.37016791118963
1.36729265920189 0.597849962429656
0.519752323523261 1.89624294678323
0.0191107582149252 5.33379579237503
0.196986770599228 4.03647401377799
0.952624784132827 -2.16148044060619
0.0015044650302265 5.2886136068202
0.902037550168227 -2.52858794998143
1.22749030238651 0.0464677514177437
1.21792105205884 -0.141830633035098
0.0102856385216306 6.14133785166822
0.863182937449097 -2.1011551265248
0.818564275101864 -1.01534185724427
0.302635244152368 1.46210778674698
0.372879666668875 1.40476340640872
0.889792662887503 -1.17727077360789
1.29641079289544 0.586535140325398
1.3272574482414 1.3843767453822
0.214325965580719 3.31884710039265
0.614137818457737 1.59184679724417
0.368321423663743 0.845629772651128
0.380933232697981 0.723731284253583
0.976159176116351 -1.93326574524374
0.857441599514585 -1.60433853663788
0.83708880960736 -2.04818343189899
1.47579759938635 -0.0805765439070799
0.442438891644387 1.73835028751105
1.38953493744193 0.417026784527823
1.14080627628873 0.550112184891199
1.48114359376883 0.298669829133318
0.736597976181201 -0.643077812882977
0.394694513831174 0.936428225891781
1.26908189826216 0.858430359609534
0.0672098890730217 5.09900003215748
0.333851467177716 3.01981294882107
0.945969964783799 -1.83629872793283
0.692843988254517 0.433249498499454
1.26598975868233 1.28216661393988
0.391972523919673 1.09693715415414
0.46917601169147 1.50263243659036
1.01098339289849 -2.31817336136178
0.777738706583271 -0.838717602177543
1.12209237980542 -0.319615415703505
1.17055379105428 0.512465979308082
0.258836084652896 2.78755665557385
0.61396534548555 1.69052927515609
1.31770140735241 1.16502121252192
0.116000397751632 4.99444756152836
0.210700698077648 3.49050340807256
1.05114996874651 -1.76278149811195
0.670454848751649 0.224613670765436
1.92406486757113 -0.108425407068489
1.94209538616793 1.7264746642509
1.99393444419407 0.403999409461401
1.95439877208446 0.827632344838344
1.98473609767015 0.0244378792655042
1.9054919129869 0.855315202370139
1.94542261961925 0.240943330223005
1.96246848086935 0.398027059259599
};
\addplot [draw=black, fill=black, forget plot, mark=|, only marks]
table{%
x  y
0.736597976181201 -5
0.332942022904396 -5
1.21589080021465 -5
0.729874319093148 -5
0.527962441193321 -5
1.02156844318979 -5
1.47579759938635 -5
1.38953493744193 -5
0.955601779278568 -5
1.36046218960706 -5
0.729180476405212 -5
0.212701803984358 -5
0.388158878064094 -5
0.484633763104316 -5
0.380933232697981 -5
0.866704965181163 -5
1.94542261961925 -5
0.130904366804652 -5
1.9054919129869 -5
0.982715141501147 -5
0.362559658595395 -5
1.92406486757113 -5
1.98473609767015 -5
0.48579647389144 -5
0.00266557165722991 -5
1.3272574482414 -5
0.0927035953462867 -5
1.94209538616793 -5
0.495989002819549 -5
0.56325131198675 -5
};
\addplot [semithick, steelblue31119180]
table {%
-0.2 6.36070232858577
-0.187939698492462 6.35472956573446
-0.175879396984925 6.34731334659466
-0.163819095477387 6.3382973339002
-0.151758793969849 6.32750597075953
-0.139698492462312 6.31474262926839
-0.127638190954774 6.29978772331929
-0.115577889447236 6.28239683324762
-0.103517587939698 6.26229890775984
-0.0914572864321608 6.23919463075493
-0.0793969849246231 6.21275506772564
-0.0673366834170854 6.18262073884089
-0.0552763819095477 6.14840130364111
-0.0432160804020101 6.10967608498086
-0.0311557788944724 6.06599570590153
-0.0190954773869347 6.01688515932343
-0.00703517587939698 5.96184867159156
0.00502512562814073 5.9003767487355
0.0170854271356784 5.83195579694493
0.0291457286432161 5.75608067051428
0.0412060301507538 5.67227040191608
0.0532663316582915 5.5800871885529
0.0653266331658292 5.4791584294262
0.0773869346733668 5.36920121117875
0.0894472361809046 5.25004814227613
0.101507537688442 5.12167286069599
0.11356783919598 4.9842129674799
0.125628140703518 4.83798768382257
0.137688442211055 4.68350735022958
0.149748743718593 4.52147215648096
0.161809045226131 4.35275835662945
0.173869346733668 4.17839174290576
0.185929648241206 3.99951023294512
0.197989949748744 3.81731977895058
0.210050251256281 3.63304995308783
0.222110552763819 3.44791688714876
0.234170854271357 3.2631011241545
0.246231155778894 3.07974591479964
0.258291457286432 2.89897745452039
0.27035175879397 2.72194293904684
0.282412060301508 2.54985622775564
0.294472361809045 2.38403613684964
0.306532663316583 2.22592111995052
0.318592964824121 2.0770481567441
0.330653266331658 1.93899336049264
0.342713567839196 1.8132847868457
0.354773869346734 1.7013090276095
0.366834170854271 1.60423608349797
0.378894472361809 1.52297756286392
0.390954773869347 1.45817288272403
0.403015075376884 1.41017517061525
0.415075376884422 1.37899584145775
0.42713567839196 1.36417616903236
0.439195979899497 1.3645893478449
0.451256281407035 1.37822749907785
0.463316582914573 1.40207037804159
0.475376884422111 1.43213649661271
0.487437185929648 1.4637669418139
0.499497487437186 1.49210269101084
0.511557788944724 1.51263277569299
0.523618090452261 1.52166051150807
0.535678391959799 1.5165737009306
0.547738693467337 1.49588586888149
0.559798994974875 1.45909325603014
0.571859296482412 1.40643323587227
0.58391959798995 1.33862886999712
0.595979899497487 1.2566771019735
0.608040201005025 1.16170473720144
0.620100502512563 1.05489024637302
0.632160804020101 0.93743495680274
0.644221105527638 0.810562848055073
0.656281407035176 0.675530347542954
0.668341708542714 0.533632875857196
0.680402010050251 0.386201076189385
0.692462311557789 0.234585255604396
0.704522613065327 0.0801307718497179
0.716582914572864 -0.0758504607712076
0.728643216080402 -0.232106540982688
0.74070351758794 -0.387465332697582
0.752763819095478 -0.540844455742336
0.764824120603015 -0.69125092017881
0.776884422110553 -0.837771945212664
0.788944723618091 -0.979559969048386
0.801005025125628 -1.11581542763761
0.813065326633166 -1.24577057390912
0.825125628140704 -1.36867662871632
0.837185929648241 -1.48379520730234
0.849246231155779 -1.59039357303285
0.861306532663317 -1.68774210323933
0.873366834170854 -1.77511158731768
0.885427135678392 -1.85176769569789
0.89748743718593 -1.91696017190033
0.909547738693467 -1.96990500045275
0.921608040201005 -2.00975902590461
0.933668341708543 -2.03558837292993
0.945728643216081 -2.04633479373056
0.957788944723618 -2.04078805457172
0.969849246231156 -2.01757782337929
0.981909547738694 -1.97520477314883
0.993969849246231 -1.91213594055423
1.00603015075377 -1.82698974108667
1.01809045226131 -1.7188250806116
1.03015075376884 -1.58752014867175
1.04221105527638 -1.43417879120378
1.05427135678392 -1.26144888283686
1.06633165829146 -1.07360863489331
1.078391959799 -0.876309780531172
1.09045226130653 -0.675973094164984
1.10251256281407 -0.47897245200267
1.11457286432161 -0.290839862148597
1.12663316582915 -0.115713508513106
1.13869346733668 0.0438591077268476
1.15075376884422 0.186790405153377
1.16281407035176 0.3131161071239
1.1748743718593 0.423594221051813
1.18693467336683 0.519345197152491
1.19899497487437 0.601584816218358
1.21105527638191 0.671457027710173
1.22311557788945 0.729949240408179
1.23517587939698 0.777864813182883
1.24723618090452 0.815829767935315
1.25929648241206 0.844316963973451
1.2713567839196 0.863677470322242
1.28341708542714 0.874174058355338
1.29547738693467 0.876015231361452
1.30753768844221 0.869390138140973
1.31959798994975 0.85450532406461
1.33165829145729 0.831623761555799
1.34371859296482 0.801105168466419
1.35577889447236 0.763444562655624
1.3678391959799 0.71930383387754
1.37989949748744 0.669529620920853
1.39195979899498 0.615150860233467
1.40402010050251 0.557351671239538
1.41608040201005 0.497419705247187
1.42814070351759 0.436675669338059
1.44020100502513 0.376394629994868
1.45226130653266 0.317732051845667
1.4643216080402 0.261666358806096
1.47638190954774 0.208965544817018
1.48844221105528 0.160179615709723
1.50050251256281 0.115655377138293
1.51256281407035 0.0755667317508104
1.52462311557789 0.0399526613831355
1.53668341708543 0.00875601553523523
1.54874371859296 -0.0181417639555125
1.5608040201005 -0.0408926322031861
1.57286432160804 -0.0596614324750676
1.58492462311558 -0.0746124102932346
1.59698492462312 -0.0859015903797752
1.60904522613065 -0.0936736256940888
1.62110552763819 -0.0980617822863846
1.63316582914573 -0.0991899227284965
1.64522613065327 -0.0971755970033192
1.6572864321608 -0.0921335863725596
1.66934673366834 -0.0841794465301677
1.68140703517588 -0.0734327541397741
1.69346733668342 -0.0600198781738963
1.70552763819095 -0.0440761812736237
1.71758793969849 -0.0257476142938168
1.72964824120603 -0.00519170617905642
1.74170854271357 0.0174220231727022
1.75376884422111 0.0419121831149164
1.76582914572864 0.0680861115785948
1.77788944723618 0.0957408938299712
1.78994974874372 0.124664738070084
1.80201005025126 0.154638644324363
1.81407035175879 0.185438303524533
1.82613065326633 0.216836165124762
1.83819095477387 0.248603613879098
1.85025125628141 0.280513199274388
1.86231155778894 0.312340864605069
1.87437185929648 0.343868126674329
1.88643216080402 0.374884161510673
1.89849246231156 0.405187756398019
1.9105527638191 0.434589093689028
1.92261306532663 0.462911337496224
1.93467336683417 0.489992000067507
1.94673366834171 0.515684070626019
1.95879396984925 0.539856895424896
1.97085427135678 0.56239680368694
1.98291457286432 0.583207479815648
1.99497487437186 0.602210087656844
2.0070351758794 0.619343157586567
2.01909547738693 0.634562251566654
2.03115577889447 0.647839425112722
2.04321608040201 0.659162508156065
2.05527638190955 0.668534229079709
2.06733668341709 0.675971207695109
2.07939698492462 0.681502843714398
2.09145728643216 0.685170127215315
2.1035175879397 0.687024397011847
2.11557788944724 0.687126071528283
2.12763819095477 0.685543375069712
2.13969849246231 0.682351080210398
2.15175879396985 0.677629284609379
2.16381909547739 0.671462237953302
2.17587939698492 0.663937232001426
2.18793969849246 0.655143564077369
2.2 0.645171581654874
};
\addlegendentry{Mean}
\addplot [semithick, red, forget plot]
table {%
-0.2 6.2857973514481
-0.187939698492462 6.28987893254762
-0.175879396984925 6.29181588372522
-0.163819095477387 6.29143317385438
-0.151758793969849 6.28854013004549
-0.139698492462312 6.28292926239899
-0.127638190954774 6.27437510005683
-0.115577889447236 6.26263307579898
-0.103517587939698 6.24743850896844
-0.0914572864321608 6.22850575195502
-0.0793969849246231 6.20552758417457
-0.0673366834170854 6.17817495967054
-0.0552763819095477 6.14609724012402
-0.0432160804020101 6.10892307376503
-0.0311557788944724 6.06626211137595
-0.0190954773869347 6.01770778129061
-0.00703517587939698 5.96284137276835
0.00502512562814073 5.90123769647965
0.0170854271356784 5.83247259523151
0.0291457286432161 5.75613255854249
0.0412060301507538 5.67182664040638
0.0532663316582915 5.5792007786667
0.0653266331658292 5.47795445565633
0.0773869346733668 5.36785941553933
0.0894472361809046 5.24877986427611
0.101507537688442 5.12069323625788
0.11356783919598 4.9837102481237
0.125628140703518 4.83809262672679
0.137688442211055 4.68426666625203
0.149748743718593 4.52283072306446
0.161809045226131 4.35455497673142
0.173869346733668 4.18037232676687
0.185929648241206 4.00136016107162
0.197989949748744 3.81871385699869
0.210050251256281 3.63371411567772
0.222110552763819 3.44769137777358
0.234170854271357 3.26199138801039
0.246231155778894 3.0779462474231
0.258291457286432 2.89685486141262
0.27035175879397 2.71997550441124
0.282412060301508 2.54853134759842
0.294472361809045 2.38372743454599
0.306532663316583 2.2267750630501
0.318592964824121 2.07891724952624
0.330653266331658 1.94144733602477
0.342713567839196 1.81571216298775
0.354773869346734 1.70309166873519
0.366834170854271 1.60494816112118
0.378894472361809 1.52254067772778
0.390954773869347 1.45690296965451
0.403015075376884 1.40868845690316
0.415075376884422 1.37799311124833
0.42713567839196 1.36417806680222
0.439195979899497 1.36572605810096
0.451256281407035 1.38017446012216
0.463316582914573 1.40416536568561
0.475376884422111 1.43363406884371
0.487437185929648 1.46412297935564
0.499497487437186 1.49116970386928
0.511557788944724 1.51069284803595
0.523618090452261 1.51929979398149
0.535678391959799 1.51446714403481
0.547738693467337 1.49458367539806
0.559798994974875 1.45888030004499
0.571859296482412 1.40729020308166
0.58391959798995 1.3402834770427
0.595979899497487 1.2587099228498
0.608040201005025 1.16366898121746
0.620100502512563 1.05641276278727
0.632160804020101 0.938279469283438
0.644221105527638 0.81065027839526
0.656281407035176 0.674921907157222
0.668341708542714 0.53248820908157
0.680402010050251 0.384726155567839
0.692462311557789 0.232983617124391
0.704522613065327 0.0785680401076337
0.716582914572864 -0.0772637885233033
0.728643216080402 -0.233314196261137
0.74070351758794 -0.388450594830113
0.752763819095478 -0.541608554626875
0.764824120603015 -0.691791981866757
0.776884422110553 -0.838070751640898
0.788944723618091 -0.979576344982615
0.801005025125628 -1.11549603458366
0.813065326633166 -1.24506598531361
0.825125628140704 -1.36756334671807
0.837185929648241 -1.48229709669328
0.849246231155779 -1.58859712629165
0.861306532663317 -1.68580089703069
0.873366834170854 -1.773236998834
0.885427135678392 -1.85020512459456
0.89748743718593 -1.915952398347
0.909547738693467 -1.96964671375817
0.921608040201005 -2.01034885974521
0.933668341708543 -2.03698686496384
0.945728643216081 -2.0483383129386
0.957788944723618 -2.04302939259167
0.969849246231156 -2.01956288769164
0.981909547738694 -1.97639030730567
0.993969849246231 -1.91204409703886
1.00603015075377 -1.82534140335192
1.01809045226131 -1.7156576560425
1.03015075376884 -1.5832440846665
1.04221105527638 -1.42953085004441
1.05427135678392 -1.25732787946063
1.06633165829146 -1.07082794450782
1.078391959799 -0.875349345818019
1.09045226130653 -0.676830101549934
1.10251256281407 -0.481174989533318
1.11457286432161 -0.293615776282351
1.12663316582915 -0.118239365629348
1.13869346733668 0.0422281830821599
1.15075376884422 0.186383598074527
1.16281407035176 0.313924893349873
1.1748743718593 0.425342584611239
1.18693467336683 0.521607339752872
1.19899497487437 0.603906758694623
1.21105527638191 0.673452590391777
1.22311557788945 0.731357540851534
1.23517587939698 0.778569491602677
1.24723618090452 0.815847807322585
1.25929648241206 0.843767921655468
1.2713567839196 0.862743743355362
1.28341708542714 0.873060917720606
1.29547738693467 0.874916788169473
1.30753768844221 0.868464737975285
1.31959798994975 0.853861424167359
1.33165829145729 0.831315338883324
1.34371859296482 0.801134330340053
1.35577889447236 0.763768482133921
1.3678391959799 0.719843528247159
1.37989949748744 0.670179340507215
1.39195979899498 0.615788538744166
1.40402010050251 0.557852295893885
1.41608040201005 0.497673827854153
1.42814070351759 0.436614167370649
1.44020100502513 0.376018470128547
1.45226130653266 0.317143106662216
1.4643216080402 0.261093458268937
1.47638190954774 0.208779774388998
1.48844221105528 0.160894532365159
1.50050251256281 0.11791068459771
1.51256281407035 0.0800970364996313
1.52462311557789 0.0475453189557888
1.53668341708543 0.0202032889778795
1.54874371859296 -0.00209098424899307
1.5608040201005 -0.0195771400316618
1.57286432160804 -0.032544923219021
1.58492462311558 -0.0413118230085809
1.59698492462312 -0.0462059543733475
1.60904522613065 -0.0475536584522557
1.62110552763819 -0.045671049320416
1.63316582914573 -0.0408586771824737
1.64522613065327 -0.0333985352335383
1.6572864321608 -0.0235527505120653
1.66934673366834 -0.0115634290159743
1.68140703517588 0.00234675048540267
1.69346733668342 0.0179734921058688
1.70552763819095 0.0351295990490659
1.71758793969849 0.0536435941307043
1.72964824120603 0.073358338275725
1.74170854271357 0.0941297045827736
1.75376884422111 0.115825340404262
1.76582914572864 0.138323532947578
1.77788944723618 0.161512182936473
1.78994974874372 0.185287884140754
1.80201005025126 0.209555102768635
1.81407035175879 0.234225448846351
1.82613065326633 0.259217031081335
1.83819095477387 0.284453886822803
1.85025125628141 0.309865479256349
1.86231155778894 0.335386254670267
1.87437185929648 0.360955253367883
1.88643216080402 0.38651576848721
1.89849246231156 0.41201504758538
1.9105527638191 0.437404032335556
1.92261306532663 0.462637132073363
1.93467336683417 0.487672027233049
1.94673366834171 0.512469498951547
1.95879396984925 0.536993281313269
1.97085427135678 0.561209932880848
1.98291457286432 0.585088724324853
1.99497487437186 0.608601539142692
2.0070351758794 0.631722784652697
2.01909547738693 0.654429310669023
2.03115577889447 0.676700333507167
2.04321608040201 0.698517363236846
2.05527638190955 0.719864132383363
2.06733668341709 0.740726524574086
2.07939698492462 0.761092501925605
2.09145728643216 0.780952030261601
2.1035175879397 0.800297001533838
2.11557788944724 0.819121153082388
2.12763819095477 0.837419983610359
2.13969849246231 0.855190665958855
2.15175879396985 0.872431956946856
2.16381909547739 0.889144104686578
2.17587939698492 0.905328753897681
2.18793969849246 0.920988849824405
2.2 0.936128541410374
};
\path [draw=steelblue31119180, fill=steelblue31119180, opacity=0.2]
(axis cs:-0.2,13.0843990932545)
--(axis cs:-0.2,-0.362994436082984)
--(axis cs:-0.187939698492462,0.188777675787692)
--(axis cs:-0.175879396984925,0.7194155291641)
--(axis cs:-0.163819095477387,1.2276342734235)
--(axis cs:-0.151758793969849,1.71201960231573)
--(axis cs:-0.139698492462312,2.17100692931417)
--(axis cs:-0.127638190954774,2.60285535575632)
--(axis cs:-0.115577889447236,3.00561466919068)
--(axis cs:-0.103517587939698,3.37708395616428)
--(axis cs:-0.0914572864321608,3.71476279813318)
--(axis cs:-0.0793969849246231,4.01580373099019)
--(axis cs:-0.0673366834170854,4.27699439199833)
--(axis cs:-0.0552763819095477,4.49483990256926)
--(axis cs:-0.0432160804020101,4.66588458829617)
--(axis cs:-0.0311557788944724,4.78746412581474)
--(axis cs:-0.0190954773869347,4.85895610538454)
--(axis cs:-0.00703517587939698,4.88310428193668)
--(axis cs:0.00502512562814073,4.8664097441523)
--(axis cs:0.0170854271356784,4.81788700769972)
--(axis cs:0.0291457286432161,4.7468098657438)
--(axis cs:0.0412060301507538,4.66080342538419)
--(axis cs:0.0532663316582915,4.56498670637322)
--(axis cs:0.0653266331658292,4.46199078967208)
--(axis cs:0.0773869346733668,4.3524267837465)
--(axis cs:0.0894472361809046,4.23553143267436)
--(axis cs:0.101507537688442,4.1098857993086)
--(axis cs:0.11356783919598,3.97415039643458)
--(axis cs:0.125628140703518,3.82771235854031)
--(axis cs:0.137688442211055,3.671078293956)
--(axis cs:0.149748743718593,3.50586273647138)
--(axis cs:0.161809045226131,3.33435345129799)
--(axis cs:0.173869346733668,3.15881865877757)
--(axis cs:0.185929648241206,2.98084968608069)
--(axis cs:0.197989949748744,2.80104290421962)
--(axis cs:0.210050251256281,2.61920934859294)
--(axis cs:0.222110552763819,2.43506718115051)
--(axis cs:0.234170854271357,2.24907625399402)
--(axis cs:0.246231155778894,2.06289705387989)
--(axis cs:0.258291457286432,1.87912249005033)
--(axis cs:0.27035175879397,1.70041682397666)
--(axis cs:0.282412060301508,1.52864242902396)
--(axis cs:0.294472361809045,1.36460492559962)
--(axis cs:0.306532663316583,1.20863893953909)
--(axis cs:0.318592964824121,1.06160751311922)
--(axis cs:0.330653266331658,0.925450349878024)
--(axis cs:0.342713567839196,0.80268485842568)
--(axis cs:0.354773869346734,0.695178970451303)
--(axis cs:0.366834170854271,0.60325203182844)
--(axis cs:0.378894472361809,0.525959291567295)
--(axis cs:0.390954773869347,0.462381152356499)
--(axis cs:0.403015075376884,0.412768584189711)
--(axis cs:0.415075376884422,0.378486097128876)
--(axis cs:0.42713567839196,0.360830583437286)
--(axis cs:0.439195979899497,0.359736475942152)
--(axis cs:0.451256281407035,0.373203945966345)
--(axis cs:0.463316582914573,0.397517886434831)
--(axis cs:0.475376884422111,0.427870348352162)
--(axis cs:0.487437185929648,0.459047567731773)
--(axis cs:0.499497487437186,0.48603415600171)
--(axis cs:0.511557788944724,0.504486543844146)
--(axis cs:0.523618090452261,0.511068716461791)
--(axis cs:0.535678391959799,0.503641981350797)
--(axis cs:0.547738693467337,0.481241611085829)
--(axis cs:0.559798994974875,0.443770490507571)
--(axis cs:0.571859296482412,0.391493149893171)
--(axis cs:0.58391959798995,0.32459934809828)
--(axis cs:0.595979899497487,0.243111228878294)
--(axis cs:0.608040201005025,0.14718878060984)
--(axis cs:0.620100502512563,0.0376058282407554)
--(axis cs:0.632160804020101,-0.0839623454128017)
--(axis cs:0.644221105527638,-0.21507465552659)
--(axis cs:0.656281407035176,-0.35303534440161)
--(axis cs:0.668341708542714,-0.495508761111208)
--(axis cs:0.680402010050251,-0.640937449897437)
--(axis cs:0.692462311557789,-0.788597038232085)
--(axis cs:0.704522613065327,-0.938306857002468)
--(axis cs:0.716582914572864,-1.08996506541067)
--(axis cs:0.728643216080402,-1.24313912800982)
--(axis cs:0.74070351758794,-1.3968899436915)
--(axis cs:0.752763819095478,-1.54986528709949)
--(axis cs:0.764824120603015,-1.7005535909524)
--(axis cs:0.776884422110553,-1.8475342532676)
--(axis cs:0.788944723618091,-1.98961250314218)
--(axis cs:0.801005025125628,-2.12582470843833)
--(axis cs:0.813065326633166,-2.25537067852785)
--(axis cs:0.825125628140704,-2.37754111489426)
--(axis cs:0.837185929648241,-2.4916758071725)
--(axis cs:0.849246231155779,-2.59714793458498)
--(axis cs:0.861306532663317,-2.69335068206714)
--(axis cs:0.873366834170854,-2.7796694996495)
--(axis cs:0.885427135678392,-2.8554420757846)
--(axis cs:0.89748743718593,-2.91992041879412)
--(axis cs:0.909547738693467,-2.97224884082878)
--(axis cs:0.921608040201005,-3.0114655804248)
--(axis cs:0.933668341708543,-3.03653463560994)
--(axis cs:0.945728643216081,-3.04641607355877)
--(axis cs:0.957788944723618,-3.04016847572654)
--(axis cs:0.969849246231156,-3.01702802622901)
--(axis cs:0.981909547738694,-2.97634859958705)
--(axis cs:0.993969849246231,-2.91731580428514)
--(axis cs:1.00603015075377,-2.83858833729743)
--(axis cs:1.01809045226131,-2.73840294100381)
--(axis cs:1.03015075376884,-2.61574432937723)
--(axis cs:1.04221105527638,-2.47238467795966)
--(axis cs:1.05427135678392,-2.3141633175627)
--(axis cs:1.06633165829146,-2.14933052630616)
--(axis cs:1.078391959799,-1.98399586159427)
--(axis cs:1.09045226130653,-1.81855293155342)
--(axis cs:1.10251256281407,-1.64895826700318)
--(axis cs:1.11457286432161,-1.47165497624636)
--(axis cs:1.12663316582915,-1.28765552788404)
--(axis cs:1.13869346733668,-1.10311864051217)
--(axis cs:1.15075376884422,-0.926902263048286)
--(axis cs:1.16281407035176,-0.767121759234727)
--(axis cs:1.1748743718593,-0.628670149281965)
--(axis cs:1.18693467336683,-0.512655157906021)
--(axis cs:1.19899497487437,-0.417436024269111)
--(axis cs:1.21105527638191,-0.340188758569525)
--(axis cs:1.22311557788945,-0.278095071365007)
--(axis cs:1.23517587939698,-0.228892834674805)
--(axis cs:1.24723618090452,-0.190974025115765)
--(axis cs:1.25929648241206,-0.163283030927265)
--(axis cs:1.2713567839196,-0.145152141021098)
--(axis cs:1.28341708542714,-0.136115250599168)
--(axis cs:1.29547738693467,-0.1357276877719)
--(axis cs:1.30753768844221,-0.143450323358098)
--(axis cs:1.31959798994975,-0.158662526014355)
--(axis cs:1.33165829145729,-0.180812232193409)
--(axis cs:1.34371859296482,-0.209610890811642)
--(axis cs:1.35577889447236,-0.245104189481458)
--(axis cs:1.3678391959799,-0.287473674701745)
--(axis cs:1.37989949748744,-0.336582207560976)
--(axis cs:1.39195979899498,-0.391500985605476)
--(axis cs:1.40402010050251,-0.450376591406479)
--(axis cs:1.41608040201005,-0.510865342404242)
--(axis cs:1.42814070351759,-0.571042746727035)
--(axis cs:1.44020100502513,-0.630438169943246)
--(axis cs:1.45226130653266,-0.690811019403245)
--(axis cs:1.4643216080402,-0.756392696372101)
--(axis cs:1.47638190954774,-0.833405708266279)
--(axis cs:1.48844221105528,-0.928761377523517)
--(axis cs:1.50050251256281,-1.04815913866131)
--(axis cs:1.51256281407035,-1.19433841953466)
--(axis cs:1.52462311557789,-1.36634152527687)
--(axis cs:1.53668341708543,-1.55996739999482)
--(axis cs:1.54874371859296,-1.76886173103594)
--(axis cs:1.5608040201005,-1.9856200455246)
--(axis cs:1.57286432160804,-2.20262655319008)
--(axis cs:1.58492462311558,-2.41261923849622)
--(axis cs:1.59698492462312,-2.60905084629251)
--(axis cs:1.60904522613065,-2.78630568166573)
--(axis cs:1.62110552763819,-2.93981052972552)
--(axis cs:1.63316582914573,-3.06606566997947)
--(axis cs:1.64522613065327,-3.16261698606273)
--(axis cs:1.6572864321608,-3.2279877852688)
--(axis cs:1.66934673366834,-3.26158660340689)
--(axis cs:1.68140703517588,-3.26360434926382)
--(axis cs:1.69346733668342,-3.23491086416673)
--(axis cs:1.70552763819095,-3.176957754025)
--(axis cs:1.71758793969849,-3.09169152453631)
--(axis cs:1.72964824120603,-2.98147879605095)
--(axis cs:1.74170854271357,-2.84904373043813)
--(axis cs:1.75376884422111,-2.69741670232317)
--(axis cs:1.76582914572864,-2.52989255372082)
--(axis cs:1.77788944723618,-2.34999627066676)
--(axis cs:1.78994974874372,-2.16145332269919)
--(axis cs:1.80201005025126,-1.96816078725203)
--(axis cs:1.81407035175879,-1.77415315117961)
--(axis cs:1.82613065326633,-1.58355261510032)
--(axis cs:1.83819095477387,-1.40048733402221)
--(axis cs:1.85025125628141,-1.22895330539789)
--(axis cs:1.86231155778894,-1.07259175423217)
--(axis cs:1.87437185929648,-0.934366581716355)
--(axis cs:1.88643216080402,-0.81617358523308)
--(axis cs:1.89849246231156,-0.718497425061827)
--(axis cs:1.9105527638191,-0.640306359796942)
--(axis cs:1.92261306532663,-0.579347116333981)
--(axis cs:1.93467336683417,-0.532838809105391)
--(axis cs:1.94673366834171,-0.498376907135968)
--(axis cs:1.95879396984925,-0.474794625999278)
--(axis cs:1.97085427135678,-0.46279222202382)
--(axis cs:1.98291457286432,-0.46521334942141)
--(axis cs:1.99497487437186,-0.48685520609221)
--(axis cs:2.0070351758794,-0.533725669174285)
--(axis cs:2.01909547738693,-0.611847284809547)
--(axis cs:2.03115577889447,-0.726021494149759)
--(axis cs:2.04321608040201,-0.879083354665598)
--(axis cs:2.05527638190955,-1.07187075418116)
--(axis cs:2.06733668341709,-1.30368951641577)
--(axis cs:2.07939698492462,-1.57289794353505)
--(axis cs:2.09145728643216,-1.87737059448627)
--(axis cs:2.1035175879397,-2.21478533498366)
--(axis cs:2.11557788944724,-2.58277250735955)
--(axis cs:2.12763819095477,-2.97898159772531)
--(axis cs:2.13969849246231,-3.40110650153895)
--(axis cs:2.15175879396985,-3.84689322387803)
--(axis cs:2.16381909547739,-4.31414185542928)
--(axis cs:2.17587939698492,-4.80070790131854)
--(axis cs:2.18793969849246,-5.30450468250762)
--(axis cs:2.2,-5.82350703229574)
--(axis cs:2.2,7.11385019560548)
--(axis cs:2.2,7.11385019560548)
--(axis cs:2.18793969849246,6.61479181066236)
--(axis cs:2.17587939698492,6.1285823653214)
--(axis cs:2.16381909547739,5.65706633133588)
--(axis cs:2.15175879396985,5.20215179309679)
--(axis cs:2.13969849246231,4.76580866195974)
--(axis cs:2.12763819095477,4.35006834786473)
--(axis cs:2.11557788944724,3.95702465041612)
--(axis cs:2.1035175879397,3.58883412900736)
--(axis cs:2.09145728643216,3.2477108489169)
--(axis cs:2.07939698492462,2.93590363096385)
--(axis cs:2.06733668341709,2.65563193180599)
--(axis cs:2.05527638190955,2.40893921234058)
--(axis cs:2.04321608040201,2.19740837097773)
--(axis cs:2.03115577889447,2.0217003443752)
--(axis cs:2.01909547738693,1.88097178794285)
--(axis cs:2.0070351758794,1.77241198434742)
--(axis cs:1.99497487437186,1.6912753814059)
--(axis cs:1.98291457286432,1.63162830905271)
--(axis cs:1.97085427135678,1.5875858293977)
--(axis cs:1.95879396984925,1.55450841684907)
--(axis cs:1.94673366834171,1.52974504838801)
--(axis cs:1.93467336683417,1.5128228092404)
--(axis cs:1.92261306532663,1.50516979132643)
--(axis cs:1.9105527638191,1.509484547175)
--(axis cs:1.89849246231156,1.52887293785787)
--(axis cs:1.88643216080402,1.56594190825443)
--(axis cs:1.87437185929648,1.62210283506501)
--(axis cs:1.86231155778894,1.69727348344231)
--(axis cs:1.85025125628141,1.78997970394667)
--(axis cs:1.83819095477387,1.8976945617804)
--(axis cs:1.82613065326633,2.01722494534984)
--(axis cs:1.81407035175879,2.14502975822867)
--(axis cs:1.80201005025126,2.27743807590076)
--(axis cs:1.78994974874372,2.41078279883936)
--(axis cs:1.77788944723618,2.5414780583267)
--(axis cs:1.76582914572864,2.66606477687801)
--(axis cs:1.75376884422111,2.781241068553)
--(axis cs:1.74170854271357,2.88388777678353)
--(axis cs:1.72964824120603,2.97109538369283)
--(axis cs:1.71758793969849,3.04019629594868)
--(axis cs:1.70552763819095,3.08880539147775)
--(axis cs:1.69346733668342,3.11487110781893)
--(axis cs:1.68140703517588,3.11673884098427)
--(axis cs:1.66934673366834,3.09322771034655)
--(axis cs:1.6572864321608,3.04372061252368)
--(axis cs:1.64522613065327,2.96826579205609)
--(axis cs:1.63316582914573,2.86768582452247)
--(axis cs:1.62110552763819,2.74368696515275)
--(axis cs:1.60904522613065,2.59895843027755)
--(axis cs:1.59698492462312,2.43724766553296)
--(axis cs:1.58492462311558,2.26339441790975)
--(axis cs:1.57286432160804,2.08330368823994)
--(axis cs:1.5608040201005,1.90383478111823)
--(axis cs:1.54874371859296,1.73257820312492)
--(axis cs:1.53668341708543,1.57747943106529)
--(axis cs:1.52462311557789,1.44624684804314)
--(axis cs:1.51256281407035,1.34547188303628)
--(axis cs:1.50050251256281,1.27946989293789)
--(axis cs:1.48844221105528,1.24912060894296)
--(axis cs:1.47638190954774,1.25133679790032)
--(axis cs:1.4643216080402,1.27972541398429)
--(axis cs:1.45226130653266,1.32627512309458)
--(axis cs:1.44020100502513,1.38322742993298)
--(axis cs:1.42814070351759,1.44439408540315)
--(axis cs:1.41608040201005,1.50570475289862)
--(axis cs:1.40402010050251,1.56507993388555)
--(axis cs:1.39195979899498,1.62180270607241)
--(axis cs:1.37989949748744,1.67564144940268)
--(axis cs:1.3678391959799,1.72608134245683)
--(axis cs:1.35577889447236,1.7719933147927)
--(axis cs:1.34371859296482,1.81182122774448)
--(axis cs:1.33165829145729,1.84405975530501)
--(axis cs:1.31959798994975,1.86767317414357)
--(axis cs:1.30753768844221,1.88223059964004)
--(axis cs:1.29547738693467,1.8877581504948)
--(axis cs:1.28341708542714,1.88446336730985)
--(axis cs:1.2713567839196,1.87250708166558)
--(axis cs:1.25929648241206,1.85191695887417)
--(axis cs:1.24723618090452,1.8226335609864)
--(axis cs:1.23517587939698,1.78462246104057)
--(axis cs:1.22311557788945,1.73799355218136)
--(axis cs:1.21105527638191,1.68310281398987)
--(axis cs:1.19899497487437,1.62060565670583)
--(axis cs:1.18693467336683,1.551345552211)
--(axis cs:1.1748743718593,1.47585859138559)
--(axis cs:1.16281407035176,1.39335397348253)
--(axis cs:1.15075376884422,1.30048307335504)
--(axis cs:1.13869346733668,1.19083685596586)
--(axis cs:1.12663316582915,1.05622851085783)
--(axis cs:1.11457286432161,0.889975251949162)
--(axis cs:1.10251256281407,0.69101336299784)
--(axis cs:1.09045226130653,0.466606743223449)
--(axis cs:1.078391959799,0.231376300531928)
--(axis cs:1.06633165829146,0.00211325651953986)
--(axis cs:1.05427135678392,-0.208734448111016)
--(axis cs:1.04221105527638,-0.395972904447898)
--(axis cs:1.03015075376884,-0.559295967966264)
--(axis cs:1.01809045226131,-0.699247220219387)
--(axis cs:1.00603015075377,-0.815391144875914)
--(axis cs:0.993969849246231,-0.906956076823322)
--(axis cs:0.981909547738694,-0.974060946710623)
--(axis cs:0.969849246231156,-1.01812762052957)
--(axis cs:0.957788944723618,-1.0414076334169)
--(axis cs:0.945728643216081,-1.04625351390235)
--(axis cs:0.933668341708543,-1.03464211024993)
--(axis cs:0.921608040201005,-1.00805247138442)
--(axis cs:0.909547738693467,-0.967561160076724)
--(axis cs:0.89748743718593,-0.913999925006536)
--(axis cs:0.885427135678392,-0.848093315611168)
--(axis cs:0.873366834170854,-0.770553674985847)
--(axis cs:0.861306532663317,-0.682133524411515)
--(axis cs:0.849246231155779,-0.583639211480716)
--(axis cs:0.837185929648241,-0.475914607432175)
--(axis cs:0.825125628140704,-0.359812142538378)
--(axis cs:0.813065326633166,-0.236170469290393)
--(axis cs:0.801005025125628,-0.10580614683688)
--(axis cs:0.788944723618091,0.0304925650454052)
--(axis cs:0.776884422110553,0.171990362842277)
--(axis cs:0.764824120603015,0.318051750594784)
--(axis cs:0.752763819095478,0.46817637561482)
--(axis cs:0.74070351758794,0.621959278296333)
--(axis cs:0.728643216080402,0.778926046044447)
--(axis cs:0.716582914572864,0.938264143868253)
--(axis cs:0.704522613065327,1.0985684007019)
--(axis cs:0.692462311557789,1.25776754944088)
--(axis cs:0.680402010050251,1.41333960227621)
--(axis cs:0.668341708542714,1.5627745128256)
--(axis cs:0.656281407035176,1.70409603948752)
--(axis cs:0.644221105527638,1.83620035163674)
--(axis cs:0.632160804020101,1.95883225901828)
--(axis cs:0.620100502512563,2.07217466450529)
--(axis cs:0.608040201005025,2.17622069379305)
--(axis cs:0.595979899497487,2.2702429750687)
--(axis cs:0.58391959798995,2.35265839189596)
--(axis cs:0.571859296482412,2.42137332185138)
--(axis cs:0.559798994974875,2.4744160215527)
--(axis cs:0.547738693467337,2.51053012667715)
--(axis cs:0.535678391959799,2.52950542051041)
--(axis cs:0.523618090452261,2.53225230655434)
--(axis cs:0.511557788944724,2.52077900754183)
--(axis cs:0.499497487437186,2.49817122601996)
--(axis cs:0.487437185929648,2.46848631589603)
--(axis cs:0.475376884422111,2.43640264487327)
--(axis cs:0.463316582914573,2.40662286964836)
--(axis cs:0.451256281407035,2.38325105218935)
--(axis cs:0.439195979899497,2.36944221974764)
--(axis cs:0.42713567839196,2.36752175462743)
--(axis cs:0.415075376884422,2.37950558578662)
--(axis cs:0.403015075376884,2.40758175704079)
--(axis cs:0.390954773869347,2.45396461309156)
--(axis cs:0.378894472361809,2.51999583416055)
--(axis cs:0.366834170854271,2.6052201351675)
--(axis cs:0.354773869346734,2.7074390847677)
--(axis cs:0.342713567839196,2.82388471526573)
--(axis cs:0.330653266331658,2.95253637110725)
--(axis cs:0.318592964824121,3.09248880036899)
--(axis cs:0.306532663316583,3.24320330036195)
--(axis cs:0.294472361809045,3.40346734809967)
--(axis cs:0.282412060301508,3.57107002648732)
--(axis cs:0.27035175879397,3.74346905411702)
--(axis cs:0.258291457286432,3.91883241899046)
--(axis cs:0.246231155778894,4.09659477571938)
--(axis cs:0.234170854271357,4.27712599431498)
--(axis cs:0.222110552763819,4.46076659314701)
--(axis cs:0.210050251256281,4.64689055758272)
--(axis cs:0.197989949748744,4.83359665368154)
--(axis cs:0.185929648241206,5.01817077980956)
--(axis cs:0.173869346733668,5.19796482703395)
--(axis cs:0.161809045226131,5.37116326196092)
--(axis cs:0.149748743718593,5.53708157649053)
--(axis cs:0.137688442211055,5.69593640650316)
--(axis cs:0.125628140703518,5.84826300910482)
--(axis cs:0.11356783919598,5.99427553852521)
--(axis cs:0.101507537688442,6.13345992208339)
--(axis cs:0.0894472361809046,6.26456485187791)
--(axis cs:0.0773869346733668,6.38597563861099)
--(axis cs:0.0653266331658292,6.49632606918033)
--(axis cs:0.0532663316582915,6.59518767073258)
--(axis cs:0.0412060301507538,6.68373737844797)
--(axis cs:0.0291457286432161,6.76535147528475)
--(axis cs:0.0170854271356784,6.84602458619014)
--(axis cs:0.00502512562814073,6.9343437533187)
--(axis cs:-0.00703517587939698,7.04059306124644)
--(axis cs:-0.0190954773869347,7.17481421326233)
--(axis cs:-0.0311557788944724,7.34452728598831)
--(axis cs:-0.0432160804020101,7.55346758166554)
--(axis cs:-0.0552763819095477,7.80196270471297)
--(axis cs:-0.0673366834170854,8.08824708568345)
--(axis cs:-0.0793969849246231,8.40970640446108)
--(axis cs:-0.0914572864321608,8.76362646337667)
--(axis cs:-0.103517587939698,9.1475138593554)
--(axis cs:-0.115577889447236,9.55917899730456)
--(axis cs:-0.127638190954774,9.99672009088227)
--(axis cs:-0.139698492462312,10.4584783292226)
--(axis cs:-0.151758793969849,10.9429923392033)
--(axis cs:-0.163819095477387,11.4489603943769)
--(axis cs:-0.175879396984925,11.9752111640252)
--(axis cs:-0.187939698492462,12.5206814556812)
--(axis cs:-0.2,13.0843990932545)
--cycle;
\addlegendimage{area legend, draw=steelblue31119180, fill=steelblue31119180, opacity=0.2}
\addlegendentry{95\% interval}

\draw (axis cs:0,-4.3) node[
  scale=0.5,
  anchor=base west,
  text=black,
  rotate=0.0
]{\inducing};
\end{axis}

\end{tikzpicture}

%% file: tables/workshop_uci_table.tex
\begin{tabular}{lccc|c|cccc|cccc}
\toprule
 & & & & & \multicolumn{4}{c|}{No $\delta$ tuning} & \multicolumn{4}{c}{$\delta$ tuning} \\
 & $N$ & $D$ & $C$ & \sc nn map & \sc bnn & \sc glm & {\sc gp} subset & \our & \sc bnn & \sc glm & {\sc gp} subset & \our \\
\midrule
\sc Australian & 690 & 14 & 2 & \val{\mathbf{0.35}}{\mathbf{0.06}} & \val{0.71}{0.03} & \val{0.43}{0.04} & \val{\mathbf{0.39}}{\mathbf{0.03}} & \val{\mathbf{0.35}}{\mathbf{0.04}} & \val{\mathbf{0.34}}{\mathbf{0.05}} & \val{\mathbf{0.35}}{\mathbf{0.05}} & \val{0.41}{0.04} & \val{\mathbf{0.35}}{\mathbf{0.04}} \\
\sc Breast cancer & 683 & 10 & 2 & \val{\mathbf{0.09}}{\mathbf{0.05}} & \val{0.72}{0.06} & \val{0.47}{0.09} & \val{0.23}{0.02} & \val{0.18}{0.02} & \val{\mathbf{0.09}}{\mathbf{0.05}} & \val{\mathbf{0.09}}{\mathbf{0.05}} & \val{\mathbf{0.13}}{\mathbf{0.03}} & \val{\mathbf{0.08}}{\mathbf{0.04}} \\
\sc Digits & 351 & 34 & 2 & \val{\mathbf{0.07}}{\mathbf{0.04}} & \val{2.35}{0.01} & \val{3.11}{0.15} & \val{1.10}{0.02} & \val{1.07}{0.03} & \val{\mathbf{0.07}}{\mathbf{0.03}} & \val{\mathbf{0.07}}{\mathbf{0.04}} & \val{0.16}{0.04} & \val{\mathbf{0.08}}{\mathbf{0.03}} \\
\sc Glass & 214 & 9 & 6 & \val{\mathbf{1.02}}{\mathbf{0.41}} & \val{1.82}{0.06} & \val{1.77}{0.07} & \val{1.14}{0.07} & \val{\mathbf{0.93}}{\mathbf{0.08}} & \val{\mathbf{0.87}}{\mathbf{0.28}} & \val{\mathbf{0.82}}{\mathbf{0.27}} & \val{1.19}{0.08} & \val{\mathbf{0.92}}{\mathbf{0.11}} \\
\sc Ionosphere & 846 & 18 & 4 & \val{\mathbf{0.38}}{\mathbf{0.05}} & \val{0.70}{0.03} & \val{\mathbf{0.37}}{\mathbf{0.04}} & \val{0.48}{0.03} & \val{\mathbf{0.39}}{\mathbf{0.03}} & \val{\mathbf{0.38}}{\mathbf{0.05}} & \val{\mathbf{0.37}}{\mathbf{0.05}} & \val{0.44}{0.03} & \val{\mathbf{0.39}}{\mathbf{0.04}} \\
\sc Satellite & 1000 & 21 & 3 & \val{\mathbf{0.24}}{\mathbf{0.02}} & \val{1.83}{0.02} & \val{0.78}{0.04} & \val{0.32}{0.01} & \val{\mathbf{0.26}}{\mathbf{0.02}} & \val{\mathbf{0.24}}{\mathbf{0.02}} & \val{\mathbf{0.24}}{\mathbf{0.02}} & \val{0.43}{0.05} & \val{0.31}{0.03} \\
\sc Vehicle & 1797 & 64 & 10 & \val{\mathbf{0.40}}{\mathbf{0.06}} & \val{1.40}{0.02} & \val{1.55}{0.01} & \val{0.88}{0.02} & \val{0.85}{0.04} & \val{\mathbf{0.38}}{\mathbf{0.06}} & \val{0.37}{0.04} & \val{0.61}{0.06} & \val{\mathbf{0.43}}{\mathbf{0.02}} \\
\sc Waveform & 6435 & 35 & 6 & \val{0.40}{0.05} & \val{1.10}{0.01} & \val{1.00}{0.02} & \val{0.44}{0.03} & \val{0.38}{0.02} & \val{\mathbf{0.35}}{\mathbf{0.04}} & \val{\mathbf{0.36}}{\mathbf{0.03}} & \val{\mathbf{0.36}}{\mathbf{0.03}} & \val{\mathbf{0.32}}{\mathbf{0.03}} \\
\bottomrule
\end{tabular}

%% file: fig/australian.tex
\begin{tikzpicture}[scale=0.5]

\definecolor{color0}{rgb}{0.274509803921569,0.509803921568627,0.705882352941177}
\definecolor{color1}{rgb}{1,0.549019607843137,0}

\begin{axis}[
height=\figureheight,
tick align=outside,
tick pos=left,
title={\sc{Australian}},
width=\figurewidth,
x grid style={white!69.0196078431373!black},
xmin=-5, xmax=105,
xtick style={color=black},
xtick={-20,0,20,40,60,80,100,120},
xticklabels={\ensuremath{-}20,0,20,40,60,80,100,120},
y grid style={white!69.0196078431373!black},
ymin=0.293618057042495, ymax=0.718102694333381,
ytick style={color=black},
ytick={0.2,0.4,0.6,0.8},
yticklabels={0.2,0.4,0.6,0.8}
]
\path [draw=color0, fill=color0, opacity=0.1]
(axis cs:1,0.551667007775037)
--(axis cs:1,0.394871369856918)
--(axis cs:2,0.356818089006717)
--(axis cs:5,0.325282883209023)
--(axis cs:10,0.315017116932386)
--(axis cs:15,0.318190775561941)
--(axis cs:20,0.315173436351568)
--(axis cs:40,0.316824771925122)
--(axis cs:60,0.313980751125033)
--(axis cs:80,0.312912813282989)
--(axis cs:100,0.315163904006588)
--(axis cs:100,0.387714527539309)
--(axis cs:100,0.387714527539309)
--(axis cs:80,0.391337690954981)
--(axis cs:60,0.38348557249058)
--(axis cs:40,0.385913440494826)
--(axis cs:20,0.38626921460719)
--(axis cs:15,0.384456222562076)
--(axis cs:10,0.384020394527795)
--(axis cs:5,0.384817186909285)
--(axis cs:2,0.422463140257074)
--(axis cs:1,0.551667007775037)
--cycle;

\path [draw=color1, fill=color1, opacity=0.1]
(axis cs:1,0.698807938092886)
--(axis cs:1,0.662510632062247)
--(axis cs:2,0.657078817702754)
--(axis cs:5,0.517882131251046)
--(axis cs:10,0.40135987267536)
--(axis cs:15,0.393191142351973)
--(axis cs:20,0.371626166919784)
--(axis cs:40,0.35299934681048)
--(axis cs:60,0.324006125784774)
--(axis cs:80,0.327733599897042)
--(axis cs:100,0.315157958626014)
--(axis cs:100,0.388343247321801)
--(axis cs:100,0.388343247321801)
--(axis cs:80,0.39632234172613)
--(axis cs:60,0.39462009153726)
--(axis cs:40,0.403204497924365)
--(axis cs:20,0.453648844608426)
--(axis cs:15,0.468843614804632)
--(axis cs:10,0.622775822605122)
--(axis cs:5,0.646032048907283)
--(axis cs:2,0.691617433808421)
--(axis cs:1,0.698807938092886)
--cycle;

\addplot [semithick, color0]
table {%
1 0.473269188815977
2 0.389640614631896
5 0.355050035059154
10 0.349518755730091
15 0.351323499062008
20 0.350721325479379
40 0.351369106209974
60 0.348733161807807
80 0.352125252118985
100 0.351439215772948
};
\addplot [semithick, color1]
table {%
1 0.680659285077567
2 0.674348125755587
5 0.581957090079165
10 0.512067847640241
15 0.431017378578302
20 0.412637505764105
40 0.378101922367423
60 0.359313108661017
80 0.362027970811586
100 0.351750602973907
};
\end{axis}

\end{tikzpicture}

%% file: fig/glass.tex
\begin{tikzpicture}[scale=0.5]

\definecolor{color0}{rgb}{0.274509803921569,0.509803921568627,0.705882352941177}
\definecolor{color1}{rgb}{1,0.549019607843137,0}

\begin{axis}[
height=\figureheight,
tick align=outside,
tick pos=left,
title={\sc{Glass}},
width=\figurewidth,
x grid style={white!69.0196078431373!black},
xmin=-5, xmax=105,
xtick style={color=black},
xtick={-20,0,20,40,60,80,100,120},
xticklabels={\ensuremath{-}20,0,20,40,60,80,100,120},
y grid style={white!69.0196078431373!black},
ymin=0.645730930143993, ymax=1.81802513016566,
ytick style={color=black}
]
\path [draw=color0, fill=color0, opacity=0.1]
(axis cs:1,1.49771547138196)
--(axis cs:1,1.18952652316133)
--(axis cs:1,1.18266373930492)
--(axis cs:5,0.96605549198983)
--(axis cs:9,0.857698870422498)
--(axis cs:15,0.78946981593607)
--(axis cs:19,0.809266568098092)
--(axis cs:39,0.762809039558936)
--(axis cs:59,0.699017030144978)
--(axis cs:79,0.729237532733376)
--(axis cs:99,0.720662924824656)
--(axis cs:99,1.04896167752313)
--(axis cs:99,1.04896167752313)
--(axis cs:79,0.995374423453727)
--(axis cs:59,1.0311643117542)
--(axis cs:39,1.02993831156299)
--(axis cs:19,1.03639284358281)
--(axis cs:15,1.04105346761967)
--(axis cs:9,1.02568275842161)
--(axis cs:5,1.16679810278829)
--(axis cs:1,1.35410828283228)
--(axis cs:1,1.49771547138196)
--cycle;

\path [draw=color1, fill=color1, opacity=0.1]
(axis cs:1,1.76473903016467)
--(axis cs:1,1.58753965664686)
--(axis cs:1,1.57139842233233)
--(axis cs:5,1.35246612424739)
--(axis cs:9,1.24713571688454)
--(axis cs:15,1.14321333399575)
--(axis cs:19,1.11838605334277)
--(axis cs:39,0.956363946082652)
--(axis cs:59,0.843593507041992)
--(axis cs:79,0.782856534900766)
--(axis cs:99,0.743711704977892)
--(axis cs:99,1.01490281645003)
--(axis cs:99,1.01490281645003)
--(axis cs:79,1.04688512529527)
--(axis cs:59,1.08499783743299)
--(axis cs:39,1.12512608223656)
--(axis cs:19,1.27061408651308)
--(axis cs:15,1.26826043220348)
--(axis cs:9,1.56711712146726)
--(axis cs:5,1.55216843320769)
--(axis cs:1,1.6887418957198)
--(axis cs:1,1.76473903016467)
--cycle;

\addplot [semithick, color0]
table {%
1 1.34362099727164
1 1.2683860110686
5 1.06642679738906
9 0.941690814422055
15 0.915261641777871
19 0.922829705840453
39 0.896373675560965
59 0.865090670949587
79 0.862305978093552
99 0.884812301173894
};
\addplot [semithick, color1]
table {%
1 1.67613934340577
1 1.63007015902607
5 1.45231727872754
9 1.4071264191759
15 1.20573688309961
19 1.19450006992792
39 1.0407450141596
59 0.96429567223749
79 0.914870830098019
99 0.879307260713959
};
\end{axis}

\end{tikzpicture}

%% file: fig/vehicle.tex
\begin{tikzpicture}[scale=0.5]

\definecolor{color0}{rgb}{0.274509803921569,0.509803921568627,0.705882352941177}
\definecolor{color1}{rgb}{1,0.549019607843137,0}

\begin{axis}[
height=\figureheight,
tick align=outside,
tick pos=left,
title={\sc{Vehicle}},
width=\figurewidth,
x grid style={white!69.0196078431373!black},
xmin=-5, xmax=105,
xtick style={color=black},
xtick={-20,0,20,40,60,80,100,120},
xticklabels={\ensuremath{-}20,0,20,40,60,80,100,120},
y grid style={white!69.0196078431373!black},
ymin=0.33658889762963, ymax=1.44889874468477,
ytick style={color=black}
]
\path [draw=color0, fill=color0, opacity=0.1]
(axis cs:1,1.15038755435856)
--(axis cs:1,0.765657401041854)
--(axis cs:2,0.553511511499976)
--(axis cs:5,0.441940300378488)
--(axis cs:10,0.411710432706833)
--(axis cs:15,0.404098865005932)
--(axis cs:20,0.408111844037873)
--(axis cs:40,0.399832731742038)
--(axis cs:60,0.417572213931155)
--(axis cs:80,0.401709211226222)
--(axis cs:100,0.395165170681198)
--(axis cs:100,0.450644493515055)
--(axis cs:100,0.450644493515055)
--(axis cs:80,0.461117677074931)
--(axis cs:60,0.456069963070282)
--(axis cs:40,0.462285134628795)
--(axis cs:20,0.458052270123755)
--(axis cs:15,0.47916822165416)
--(axis cs:10,0.484698373078672)
--(axis cs:5,0.498351086862097)
--(axis cs:2,0.602037358367972)
--(axis cs:1,1.15038755435856)
--cycle;

\path [draw=color1, fill=color1, opacity=0.1]
(axis cs:1,1.39833920618226)
--(axis cs:1,1.22376840468342)
--(axis cs:2,1.10323118544521)
--(axis cs:5,0.859922498682341)
--(axis cs:10,0.674892426917935)
--(axis cs:15,0.61186036312045)
--(axis cs:20,0.557282988926062)
--(axis cs:40,0.454300742455272)
--(axis cs:60,0.438967473417847)
--(axis cs:80,0.387148436132136)
--(axis cs:100,0.394345487659343)
--(axis cs:100,0.45273904142386)
--(axis cs:100,0.45273904142386)
--(axis cs:80,0.473040831585506)
--(axis cs:60,0.468183889569496)
--(axis cs:40,0.532897960515946)
--(axis cs:20,0.668228178576085)
--(axis cs:15,0.642354417900603)
--(axis cs:10,0.789313612445954)
--(axis cs:5,1.32421902047439)
--(axis cs:2,1.18803037416823)
--(axis cs:1,1.39833920618226)
--cycle;

\addplot [semithick, color0]
table {%
1 0.958022477700206
2 0.577774434933974
5 0.470145693620292
10 0.448204402892752
15 0.441633543330046
20 0.433082057080814
40 0.431058933185416
60 0.436821088500718
80 0.431413444150576
100 0.422904832098126
};
\addplot [semithick, color1]
table {%
1 1.31105380543284
2 1.14563077980672
5 1.09207075957837
10 0.732103019681944
15 0.627107390510526
20 0.612755583751074
40 0.493599351485609
60 0.453575681493672
80 0.430094633858821
100 0.423542264541601
};
\end{axis}

\end{tikzpicture}

%% file: fig/waveform.tex
\begin{tikzpicture}[scale=0.5]

\definecolor{color0}{rgb}{0.274509803921569,0.509803921568627,0.705882352941177}
\definecolor{color1}{rgb}{1,0.549019607843137,0}

\begin{axis}[
height=\figureheight,
tick align=outside,
tick pos=left,
title={\sc{Waveform}},
width=\figurewidth,
x grid style={white!69.0196078431373!black},
xmin=-5, xmax=105,
xtick style={color=black},
xtick={-20,0,20,40,60,80,100,120},
xticklabels={\ensuremath{-}20,0,20,40,60,80,100,120},
y grid style={white!69.0196078431373!black},
ymin=0.25101727289653, ymax=1.10562919367896,
ytick style={color=black}
]
\path [draw=color0, fill=color0, opacity=0.1]
(axis cs:1,0.361220249945887)
--(axis cs:1,0.308460916257588)
--(axis cs:2,0.29482063542337)
--(axis cs:5,0.293975232230928)
--(axis cs:10,0.292210999709874)
--(axis cs:15,0.290093994205879)
--(axis cs:20,0.290933342136901)
--(axis cs:40,0.289863269295731)
--(axis cs:60,0.292746649847028)
--(axis cs:80,0.290787993278927)
--(axis cs:100,0.293411324836065)
--(axis cs:100,0.354698022519979)
--(axis cs:100,0.354698022519979)
--(axis cs:80,0.355180862488072)
--(axis cs:60,0.355566166874829)
--(axis cs:40,0.357476578371984)
--(axis cs:20,0.355392981716428)
--(axis cs:15,0.355636350111085)
--(axis cs:10,0.354116561715074)
--(axis cs:5,0.357672329791119)
--(axis cs:2,0.358965881376987)
--(axis cs:1,0.361220249945887)
--cycle;

\path [draw=color1, fill=color1, opacity=0.1]
(axis cs:1,1.06678319727976)
--(axis cs:1,0.865426190882068)
--(axis cs:2,0.565319244073352)
--(axis cs:5,0.371873220014745)
--(axis cs:10,0.414777291836851)
--(axis cs:15,0.350987260812358)
--(axis cs:20,0.329367246585807)
--(axis cs:40,0.303834266197827)
--(axis cs:60,0.291538182173199)
--(axis cs:80,0.291985301459517)
--(axis cs:100,0.291604159166322)
--(axis cs:100,0.355692878110783)
--(axis cs:100,0.355692878110783)
--(axis cs:80,0.360045802158128)
--(axis cs:60,0.357548172519876)
--(axis cs:40,0.364224010927736)
--(axis cs:20,0.390546623344611)
--(axis cs:15,0.406220806584279)
--(axis cs:10,0.442222929453093)
--(axis cs:5,0.849592195755218)
--(axis cs:2,0.918583735910069)
--(axis cs:1,1.06678319727976)
--cycle;

\addplot [semithick, color0]
table {%
1 0.334840583101738
2 0.326893258400179
5 0.325823781011023
10 0.323163780712474
15 0.322865172158482
20 0.323163161926664
40 0.323669923833858
60 0.324156408360928
80 0.322984427883499
100 0.324054673678022
};
\addplot [semithick, color1]
table {%
1 0.966104694080912
2 0.741951489991711
5 0.610732707884982
10 0.428500110644972
15 0.378604033698319
20 0.359956934965209
40 0.334029138562782
60 0.324543177346538
80 0.326015551808822
100 0.323648518638553
};
\end{axis}

\end{tikzpicture}

%% file: fig/digits.tex
\begin{tikzpicture}[scale=0.5]

\definecolor{color0}{rgb}{0.274509803921569,0.509803921568627,0.705882352941177}
\definecolor{color1}{rgb}{1,0.549019607843137,0}

\begin{axis}[
height=\figureheight,
tick align=outside,
tick pos=left,
title={\sc{Digits}},
width=\figurewidth,
x grid style={white!69.0196078431373!black},
xmin=-5, xmax=105,
xtick style={color=black},
xtick={-20,0,20,40,60,80,100,120},
xticklabels={\ensuremath{-}20,0,20,40,60,80,100,120},
y grid style={white!69.0196078431373!black},
ymin=-0.0757482569830465, ymax=2.64986615846764,
ytick style={color=black},
ytick={-1,0,1,2,3},
yticklabels={\ensuremath{-}1,0,1,2,3}
]
\path [draw=color0, fill=color0, opacity=0.1]
(axis cs:1,0.39323836656894)
--(axis cs:1,0.256974048071687)
--(axis cs:2,0.110436886108037)
--(axis cs:5,0.0701129651463406)
--(axis cs:10,0.0546518263639399)
--(axis cs:15,0.0560351102542385)
--(axis cs:20,0.051508710374841)
--(axis cs:40,0.0481433073556211)
--(axis cs:60,0.0490517206206417)
--(axis cs:80,0.0485608454909576)
--(axis cs:100,0.0488338301716529)
--(axis cs:100,0.102153097840172)
--(axis cs:100,0.102153097840172)
--(axis cs:80,0.10759918339464)
--(axis cs:60,0.108486970522196)
--(axis cs:40,0.11048702400972)
--(axis cs:20,0.108655379236024)
--(axis cs:15,0.109115547127712)
--(axis cs:10,0.117265124864633)
--(axis cs:5,0.124850625598494)
--(axis cs:2,0.179641531983892)
--(axis cs:1,0.39323836656894)
--cycle;

\path [draw=color1, fill=color1, opacity=0.1]
(axis cs:1,2.52597459412897)
--(axis cs:1,1.67096648669258)
--(axis cs:2,1.05819094956436)
--(axis cs:5,0.436078124387862)
--(axis cs:10,0.246570375717354)
--(axis cs:15,0.191218363744933)
--(axis cs:20,0.12276288402079)
--(axis cs:40,0.0743843975592598)
--(axis cs:60,0.0583990268605711)
--(axis cs:80,0.0518397263412608)
--(axis cs:100,0.0491905669827712)
--(axis cs:100,0.110905160737147)
--(axis cs:100,0.110905160737147)
--(axis cs:80,0.112012725016726)
--(axis cs:60,0.11817535751282)
--(axis cs:40,0.131702285308619)
--(axis cs:20,0.19603394219404)
--(axis cs:15,0.221348896317045)
--(axis cs:10,0.408500790578937)
--(axis cs:5,0.521238214780209)
--(axis cs:2,2.47805106813564)
--(axis cs:1,2.52597459412897)
--cycle;

\addplot [semithick, color0]
table {%
1 0.325106207320313
2 0.145039209045965
5 0.0974817953724175
10 0.0859584756142866
15 0.0825753286909755
20 0.0800820448054327
40 0.0793151656826707
60 0.0787693455714187
80 0.0780800144427987
100 0.0754934640059124
};
\addplot [semithick, color1]
table {%
1 2.09847054041078
2 1.76812100885
5 0.478658169584036
10 0.327535583148146
15 0.206283630030989
20 0.159398413107415
40 0.103043341433939
60 0.0882871921866955
80 0.0819262256789933
100 0.0800478638599592
};
\end{axis}

\end{tikzpicture}

%% file: fig/satellite.tex
\begin{tikzpicture}[scale=0.5]

\definecolor{color0}{rgb}{0.274509803921569,0.509803921568627,0.705882352941177}
\definecolor{color1}{rgb}{1,0.549019607843137,0}

\begin{axis}[
height=\figureheight,
tick align=outside,
tick pos=left,
title={\sc{Satellite}},
width=\figurewidth,
x grid style={white!69.0196078431373!black},
xmin=-5, xmax=105,
xtick style={color=black},
xtick={-20,0,20,40,60,80,100,120},
xticklabels={\ensuremath{-}20,0,20,40,60,80,100,120},
y grid style={white!69.0196078431373!black},
ymin=0.190819314356896, ymax=2.10730616991387,
ytick style={color=black}
]
\path [draw=color0, fill=color0, opacity=0.1]
(axis cs:1,0.374222729923406)
--(axis cs:1,0.361269936747104)
--(axis cs:2,0.301698098022718)
--(axis cs:5,0.294677525878294)
--(axis cs:10,0.28339422558281)
--(axis cs:15,0.277932353245849)
--(axis cs:20,0.282093962420076)
--(axis cs:40,0.28277575963665)
--(axis cs:60,0.285476892734014)
--(axis cs:80,0.292334145328354)
--(axis cs:80,0.317930971729849)
--(axis cs:80,0.317930971729849)
--(axis cs:60,0.315927804239872)
--(axis cs:40,0.323976125522112)
--(axis cs:20,0.333983045706779)
--(axis cs:15,0.347965655224169)
--(axis cs:10,0.326319373553662)
--(axis cs:5,0.334998898479628)
--(axis cs:2,0.344700051640933)
--(axis cs:1,0.374222729923406)
--cycle;

\path [draw=color1, fill=color1, opacity=0.1]
(axis cs:1,2.02019313102491)
--(axis cs:1,1.27222628709328)
--(axis cs:2,0.690396778039753)
--(axis cs:5,0.699053422642779)
--(axis cs:10,0.546633171728452)
--(axis cs:15,0.439765875514)
--(axis cs:20,0.381328178764776)
--(axis cs:40,0.35343103268326)
--(axis cs:60,0.321751924904205)
--(axis cs:80,0.295869457933797)
--(axis cs:80,0.385032093461724)
--(axis cs:80,0.385032093461724)
--(axis cs:60,0.377495936726356)
--(axis cs:40,0.433169803696785)
--(axis cs:20,0.471362444572551)
--(axis cs:15,0.608593766418142)
--(axis cs:10,0.667648413556866)
--(axis cs:5,0.746105546974112)
--(axis cs:2,1.58611908848435)
--(axis cs:1,2.02019313102491)
--cycle;

\addplot [semithick, color0]
table {%
1 0.367746333335255
2 0.323199074831825
5 0.314838212178961
10 0.304856799568236
15 0.312949004235009
20 0.308038504063427
40 0.303375942579381
60 0.300702348486943
80 0.305132558529102
100 0.286232739258317
};
\addplot [semithick, color1]
table {%
1 1.6462097090591
2 1.13825793326205
5 0.722579484808445
10 0.607140792642659
15 0.524179820966071
20 0.426345311668664
40 0.393300418190023
60 0.34962393081528
80 0.34045077569776
};
\end{axis}

\end{tikzpicture}